\documentclass[journal]{IEEEtran}
\ifCLASSINFOpdf
  \usepackage[pdftex]{graphicx}
\else
\fi
\usepackage[cmex10]{amsmath}
\usepackage{amssymb}
\usepackage{multirow}
\usepackage{booktabs}
\usepackage[nocompress]{cite}

\usepackage{xcolor}

\usepackage[colorlinks=true,linkcolor=blue,citecolor=blue,urlcolor=blue,pdfborder={0 0 0}]{hyperref}

\makeatletter
\AtBeginDocument{%
}
\makeatother

\begin{document}
%
\title{Parameter-Efficient Modality-Balanced Symmetric Fusion for Multimodal Remote Sensing Semantic Segmentation}
%
%
%

\author{Haocheng Li,
        Juepeng Zheng,~\IEEEmembership{Member,~IEEE},
        Shuangxi Miao,
        Ruibo Lu,
        Guosheng Cai,\\
        Haohuan Fu,~\IEEEmembership{Fellow,~IEEE},
        and~Jianxi Huang,~\IEEEmembership{Senior Member,~IEEE}

\thanks{This work was supported in part by the State Key Program of the National Natural Science Foundation of China under Grant 42530109 and in part by the Henan Provincial Natural Science Foundation under Grant 252300423933. \textit{(Corresponding authors: Juepeng Zheng and Jianxi Huang.)}}

\thanks{Haocheng Li, Shuangxi Miao, and Jianxi Huang are with the College of Land Science and Technology, China Agricultural University, and also with the Key Laboratory of Remote Sensing for Agri-Hazards, Ministry of Agriculture and Rural Affairs, Beijing 100083, China. Jianxi Huang is also with the Faculty of Geosciences and Engineering, Southwest Jiaotong University, Chengdu 611756, China (e-mail: haoc\_li@cau.edu.cn; jxhuang@cau.edu.cn).}
\thanks{Juepeng Zheng is with the School of Artificial Intelligence, Sun Yat-Sen University, Zhuhai, China (e-mail: zhengjp8@mail.sysu.edu.cn).}

\thanks{Ruibo Lu and Guosheng Cai are with Henan Polytechnic University, and also with the Key Laboratory of Spatio-Temporal Information and Ecological Restoration of Mines, Ministry of Natural Resources of the People’s Republic of China, Jiaozuo 454003, China.}

\thanks{Haohuan Fu is with the Tsinghua Shenzhen International Graduate School, Tsinghua University, Shenzhen, China, also with the National Supercomputing Center in Shenzhen, Shenzhen, China, and also with the Ministry of Education Key Laboratory for Earth System Modeling and the Department of Earth System Science, Tsinghua University, Beijing, China (e-mail: haohuan@tsinghua.edu.cn).}

}

%
%

\markboth{Journal of \LaTeX\ Class Files,~Vol.~13, No.~9, September~2014}%
{Shell \MakeLowercase{\textit{et al.}}: Bare Demo of IEEEtran.cls for Journals}
%



\maketitle

\begin{abstract}
Multimodal remote sensing semantic segmentation enhances scene interpretation by exploiting complementary physical cues from heterogeneous data. Although pretrained Vision Foundation Models (VFMs) provide strong general-purpose representations, adapting them to multimodal tasks often incurs substantial computational overhead and is prone to modality imbalance, where the contribution of auxiliary modalities is suppressed during optimization. To address these challenges, we propose MoBaNet, a parameter-efficient and modality-balanced symmetric fusion framework. Built upon a largely frozen VFM backbone, MoBaNet adopts a symmetric dual-stream architecture to preserve generalizable representations while minimizing the number of trainable parameters. Specifically, we design a Cross-modal Prompt-Injected Adapter (CPIA) to enable deep semantic interaction by generating shared prompts and injecting them into bottleneck adapters under the frozen backbone. To obtain compact and discriminative multimodal representations for decoding, we further introduce a Difference-Guided Gated Fusion Module (DGFM), which adaptively fuses paired stage features by explicitly leveraging cross-modal discrepancy to guide feature selection. Furthermore, we propose a Modality-Conditional Random Masking (MCRM) strategy to mitigate modality imbalance by masking one modality only during training and imposing hard-pixel auxiliary supervision on modality-specific branches. Extensive experiments on the ISPRS Vaihingen and Potsdam benchmarks demonstrate that MoBaNet achieves state-of-the-art performance with significantly fewer trainable parameters than full fine-tuning, validating its effectiveness for robust and balanced multimodal fusion. The source code in this work is available at \url{https://github.com/sauryeo/MoBaNet}.
\end{abstract}


%
\IEEEpeerreviewmaketitle

\section{Introduction}
\IEEEPARstart{R}{emote} sensing semantic segmentation aims to assign a semantic label to each pixel in high-resolution Earth observation (EO) imagery~\cite{zhu_deep_2017}, serving as a fundamental building block for a wide range of applications, including land-cover mapping, urban monitoring, precision agriculture, and disaster assessment~\cite{diakogiannis_resunet-_2020,chen_adaptive_2021,turkoglu_crop_2021,zheng_building_2021}. Compared with semantic segmentation in natural scenes, RS imagery typically exhibits larger scale variations, more frequent occlusions, and more complex backgrounds~\cite{maggiori_high-resolution_2017,volpi_dense_2017,liu_semantic_2018}. Moreover, EO data are highly susceptible to illumination changes, cast shadows, seasonal variations, and sensor-induced imaging artifacts~\cite{zhang_object-oriented_2014,bai_domain_2022}. These factors often lead to pronounced appearance ambiguity and inter-class confusion, posing more stringent requirements on the robustness of dense prediction models~\cite{ma_deep_2019}. To mitigate these challenges, researchers have increasingly incorporated complementary sensing modalities beyond RGB~\cite{schmitt_data_2016,li_deep_2022}. By fusing the physical cues encoded in heterogeneous modalities, multimodal approaches can compensate for the observational limitations of optical imagery alone in complex environments, thereby improving the completeness and stability of scene interpretation~\cite{audebert_beyond_2018,yang_attention-fused_2021}.

Although multimodal data provide rich complementary cues for RS semantic segmentation, it remains challenging to fuse heterogeneous modalities efficiently and reliably within a unified framework~\cite{parmehr_automatic_2014,sun_fully_2018}. On the one hand, different modalities exhibit substantial discrepancies in imaging mechanisms, numerical distributions, and noise characteristics—often exacerbated by resolution mismatch and registration errors—which complicate cross-modal feature alignment and interaction modeling~\cite{zheng_gather--guide_2022}. On the other hand, existing fusion strategies tend to be dominated by RGB information during optimization, such that the effective contribution of auxiliary modalities is easily suppressed and complementary cues cannot be consistently translated into performance gains~\cite{zheng_reducing_nodate,wei_improving_nodate}. In recent years, Vision Transformers (ViTs)~\cite{dosovitskiy_image_2021} and vision foundation models (VFMs)~\cite{oquab_dinov2_nodate,kirillov_segment_2023} pretrained on large-scale data have demonstrated stronger general-purpose representations, offering new opportunities for RS semantic segmentation. However, in multimodal settings, mainstream approaches typically instantiate separate backbones for different modalities and perform full fine-tuning~\cite{wang_self-supervised_2022,li_mcanet_2022}, which incurs substantial training and storage overhead. More importantly, such designs still struggle to prevent optimization from converging to RGB-dominant shortcut solutions, suggesting that architectural symmetry does not necessarily yield balanced contributions across modalities~\cite{zheng_reducing_nodate,wei_improving_nodate,niu_hybrid_2022}.

To address the above issues, we propose MoBaNet (Modality-Balanced Symmetric Fusion),
a parameter-efficient symmetric fusion framework for multimodal remote sensing semantic segmentation. 
MoBaNet adopts a pretrained vision foundation model as a shared feature extraction backbone and freezes the backbone parameters to the greatest extent 
so as to preserve its general-purpose representations~\cite{oquab_dinov2_nodate,kirillov_segment_2023}. Meanwhile, it introduces only a small set of lightweight trainable modules to model cross-modal interactions. In addition, a symmetric dual-stream design is employed to explicitly enforce equal status of different modalities during fusion, thereby mitigating the RGB-dominant bias commonly observed in multimodal optimization.

Specifically, to overcome the limited deep cross-modal interaction caused by freezing the backbone, we first design a Cross-modal Prompt-Injected Adapter (CPIA). Without disrupting the pretrained feature space of the foundation model, CPIA generates shared semantic prompts from both modalities and injects them into the bottleneck of lightweight adapters, enabling implicit semantic-level interaction and modulation across modalities~\cite{houlsby_parameter-efficient_nodate}.
Second, to obtain compact and discriminative multimodal representations for decoding, we propose a Difference-Guided Gated Fusion Module (DGFM). DGFM explicitly exploits cross-modal discrepancy to perform channel-wise and spatially adaptive fusion on paired stage features, thereby enhancing complementary information integration in a lightweight manner. This design not only suppresses the interference of noisy or inconsistent responses during fusion, but also yields more robust multimodal representations with low computational overhead.
Finally, to alleviate the tendency of multimodal models to exhibit RGB-dominant bias during training, we introduce a Modality-Conditional Random Masking (MCRM) strategy. MCRM randomly applies local masking to only one modality at the batch level and imposes hard-pixel auxiliary supervision on modality-specific branches under partial modality degradation. This mechanism encourages the model to exploit informative cues from the complementary modality when the dominant modality becomes unreliable, and thus improves the robustness and balance of cross-modal representation learning. The main contributions of this paper are as follows:

\begin{itemize}

\item We propose MoBaNet, a parameter-efficient modality-balanced symmetric fusion framework built upon VFMs for multimodal remote sensing semantic segmentation. By combining a symmetric dual-stream design with PEFT under a largely frozen backbone, MoBaNet substantially reduces training overhead while preserving the VFM's general-purpose representations and promoting balanced contributions across heterogeneous modalities.

\item We design the CPIA and DGFM modules. CPIA enables deep semantic interaction under a frozen backbone via prompt injection into bottleneck adapters, whereas DGFM performs stage-level adaptive fusion by explicitly leveraging cross-modal discrepancy to produce compact and discriminative multimodal representations for decoding.

\item We introduce an MCRM strategy. By applying modality-conditional random masking and hard-pixel auxiliary supervision during training, MCRM mitigates optical-dominant bias in multimodal fusion and improves the robustness and balance of cross-modal representation learning in complex scenes.

\item Extensive experiments on multiple multimodal remote sensing semantic segmentation benchmarks demonstrate that our method achieves state-of-the-art performance with only 6.18M trainable parameters, highlighting its strong effectiveness under a parameter-efficient adaptation setting.
\end{itemize}

The remainder of this paper is organized as follows. Section~\ref{sec:related} reviews related work. Section~\ref{sec:method} describes the proposed method. Section~\ref{sec:exp} presents the experimental setup, reports the main results, and conducts ablation studies. Section~\ref{sec:discussion} provides further discussion and analysis. Finally, Section~\ref{sec:conclusion} concludes this paper.

\section{Related Work}\label{sec:related}
\subsection{Multimodal Fusion for Remote Sensing Segmentation}
Multimodal remote sensing semantic segmentation improves class separability and robustness by fusing complementary sensing cues, e.g., multispectral appearance, DSM-derived elevation, and SAR scattering characteristics~\cite{zhu_deep_2017,li_deep_2022}. Existing methods typically adopt dual- or multi-branch encoders to extract modality-specific features, and can be broadly grouped into three paradigms according to the fusion stage: \emph{early fusion} at the input level, \emph{late fusion} at the prediction/decision level, and \emph{feature-level fusion} within the encoder--decoder pipeline~\cite{baltrusaitis_multimodal_2019}. Among them, feature-level fusion has become a predominant direction because it enables explicit cross-modal interaction and semantic alignment while preserving modality-specific characteristics~\cite{li_deep_2022,liu_transformer-based_2024}.

Building upon feature-level fusion, recent research has evolved from parallel architectures toward interaction-enhanced designs, aimed at boosting the effective contribution of auxiliary modalities while suppressing redundant information~\cite{li_deep_2022}. These methodologies can be broadly categorized into three mechanisms: (i) \textit{Cross-modal guidance}, which constrains the decoding process of one modality using structural or semantic cues from another (e.g., Gather-to-Guide~\cite{zheng_gather--guide_2022} ensures consistency between auxiliary cues and semantic decoding); (ii) \textit{Selective fusion}, utilizing gating mechanisms or attention weights to attenuate modality-specific noise and highlight complementary regions~\cite{hosseinpour_cmgfnet_2022}, as exemplified by CEGFNet's~\cite{zhou_cegfnet_2022} explicit gating for redundancy reduction; and (iii) \textit{Intra-/inter-modality joint interaction}, which establishes explicit cross-modal interaction pathways while maintaining modality-specific representations. A representative work is MFNet~\cite{ma_unified_2025}, which integrates LiDAR-derived features with image cues via meticulously designed intra- and inter-modality interaction paths~\cite{ma_crossmodal_2022}.

Meanwhile, the emergence of datasets and benchmarks tailored for heterogeneous modalities has also facilitated the development of multimodal remote sensing segmentation methods~\cite{schmitt_data_2016,liu_cross-city_2024}. For example, WHU-OPT-SAR and its accompanying MCANet~\cite{li_mcanet_2022} further validated the complementary benefits of heterogeneous modalities through cross-modal attention and multi-level feature fusion. Toward stronger interaction modeling, Transformer-based fusion has gradually become an important direction. FTransUNet~\cite{ma_multilevel_2024} introduces a fusion Transformer to perform global interaction over high-level semantics on top of a CNN backbone that captures shallow details, whereas TMFNet~\cite{liu_transformer-based_2024} and related approaches further explore DSM-oriented attention interactions and correlation modeling to enhance long-range dependency modeling and cross-modal semantic coupling. Overall, existing multimodal RS segmentation methods have progressively evolved from fusion-level design to stronger interaction modeling, with increasing emphasis on selective noise suppression/alignment and global context modeling~\cite{li_semantic_2026}.

\subsection{Foundation Models and Parameter-Efficient Adaptation}
Unlike traditional CNN-based architectures, VFMs are largely built upon ViTs and learn robust visual representations via self-supervised or weakly supervised pre-training. Among them, SAM~\cite{kirillov_segment_2023} is centered on \textit{promptable segmentation}, encapsulating segmentation as a prompt-driven universal interface and showing competitive cross-domain transferability. DINOv2~\cite{oquab_dinov2_nodate}, in contrast, follows a self-supervised representation learning paradigm, acquiring robust patch- and image-level features and serving as a general-purpose backbone for downstream tasks such as classification, detection, and segmentation. Meanwhile, the remote sensing community has also begun exploring foundation models tailored for multimodal data, such as RingMo-SAM~\cite{yan_ringmo-sam_2023} with a dedicated prompt encoder and class-decoupled mask decoder, and SkySense~\cite{guo_skysense_2024} with multimodal spatiotemporal encoding for joint cross-modal modeling. Overall, foundation models offer strong transferability and data-efficient adaptation for multimodal RS understanding.

However, the parameter scale and training cost of foundation models substantially raise the barrier to full fine-tuning. This has stimulated growing interest in PEFT. The central idea of PEFT is to keep the backbone largely frozen and introduce only a small set of trainable modules for task- or domain-specific adaptation. Representative strategies include Adapters~\cite{houlsby_parameter-efficient_nodate}, which insert lightweight bottleneck modules between layers; LoRA~\cite{hu_lora_2022}, which injects low-rank trainable updates into linear projections; and visual prompt tuning (VPT)~\cite{avidan_visual_2022}, which learns a small number of prompt vectors in the input token space. Prior studies have systematically reviewed PEFT methods and broadly categorized them into additive, reparameterization-based, and prompt-based schemes. Overall, PEFT can achieve competitive performance under a limited trainable-parameter budget while substantially reducing training and storage overhead. In remote sensing, PEFT is often used with foundation models to bridge the gap between general-purpose visual representations and domain-specific texture statistics and semantic requirements~\cite{hu_airs_2024,hu_tea_2024}.

Most existing PEFT methods are primarily designed for single-modality vision tasks, and directly transferring them to multimodal remote sensing scenarios faces critical bottlenecks~\cite{hu_airs_2024,zou_adapting_2025}. First, deep cross-modal interaction is insufficient: in mainstream dual-stream architectures, if adapters are configured independently for each modality, the frozen backbone branches lack adequate channels for deep information exchange. As a result, heterogeneous modalities are processed in a manner closer to parallel feature extraction with weakly coupled fusion, making it difficult to achieve deep complementary coupling during encoding~\cite{ma_multilevel_2024}. Second, modality imbalance tends to be reinforced under PEFT: frozen foundation models often inherit a preference for optical cues, and standard PEFT does not explicitly model the competition between modalities~\cite{wang_imbalance_2023,chen_novel_2024}. Consequently, the limited trainable parameters may still favor optical-dominant shortcut solutions, further diminishing the effective contribution of the auxiliary modality~\cite{gao_global_2024}. Therefore, it remains an open and important problem for multimodal RS semantic segmentation to develop a unified adaptation framework that, while maintaining parameter efficiency, can break modality isolation, enable deep interaction, and promote balanced contributions across modalities~\cite{hong_more_2021}.

\begin{figure*}[t]
\centering
\includegraphics[width=\linewidth]{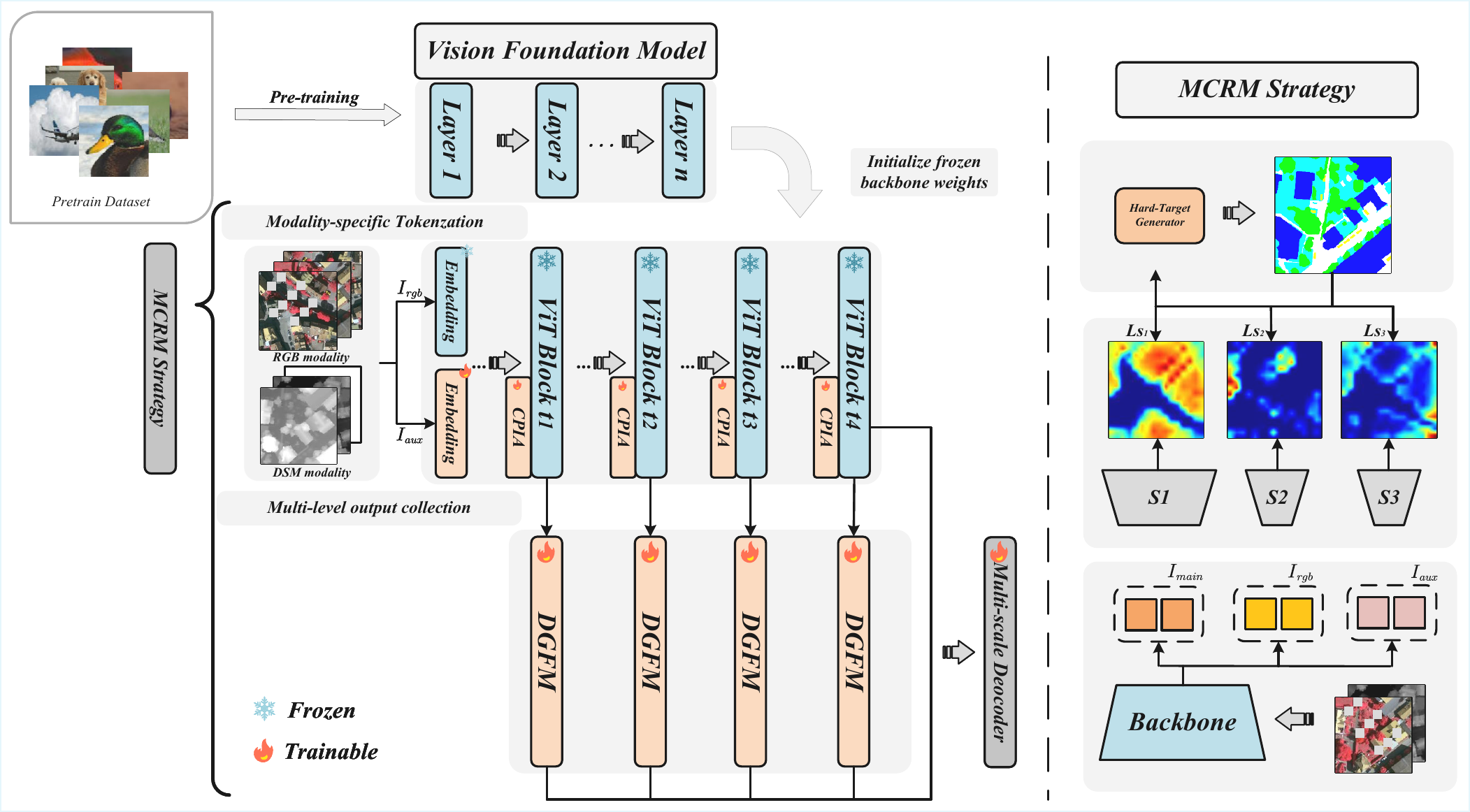}
\caption{Overall framework of the proposed multimodal segmentation model with MCRM training. 
A pretrained vision foundation model initializes a shared frozen ViT backbone. RGB and DSM inputs are first tokenized by modality-specific embeddings, then processed by a stage-wise dual-stream encoder, where CPIA is inserted before each selected stage and DGFM is applied after each selected stage to produce multi-level fused features for the decoder. During training, MCRM performs modality-conditional random masking and hard-pixel auxiliary supervision on one main branch (S1) and two auxiliary modality branches (S2, S3), while only the main branch is used for inference.}
\label{fig:overall}
\end{figure*}

\section{Method}\label{sec:method}
In this section, we first introduce the overall framework in Section~\ref{sec:overall}. The details of CPIA, DGFM, and MCRM are presented in Sections~\ref{sec:cpia}, ~\ref{sec:DGFM}, and ~\ref{sec:mcrm}, respectively.

\subsection{Overall Architecture}\label{sec:overall}
To achieve parameter-efficient and modality-balanced segmentation while preserving the representation capacity of a VFM, we propose a modality-symmetric fine-tuning framework with a dual-stream architecture. Given an input multimodal image pair consisting of an optical image $\mathbf{I}_{\mathrm{rgb}} \in \mathbb{R}^{B \times 3 \times H \times W}$ and an auxiliary modality $\mathbf{I}_{\mathrm{aux}} \in \mathbb{R}^{B \times C_a \times H \times W}$, our goal is to predict a pixel-wise class probability map $\mathbf{P} \in \mathbb{R}^{B \times K \times H \times W}$.

We adopt a pretrained ViT as the feature extraction backbone and freeze its parameters $\Theta_{\text{backbone}}$ during training. To accommodate the statistical distribution shift and channel mismatch of heterogeneous auxiliary modalities, we design an asymmetric patch embedding scheme: the optical branch reuses and freezes the pretrained embedding $E_{\mathrm{rgb}}$, whereas the auxiliary branch introduces an additional embedding $E_{\mathrm{aux}}$ whose parameters are trainable. After patch embedding, both modalities are represented as a 2-D token grid:
\begin{align}
\mathbf{X}_0 &= E_{\mathrm{rgb}}\!\left(\mathbf{I}_{\mathrm{rgb}}\right), \quad 
\mathbf{Y}_0 = E_{\mathrm{aux}}\!\left(\mathbf{I}_{\mathrm{aux}}\right), \label{eq:patch_embed} \\
\mathbf{X}_0, \mathbf{Y}_0 &\in \mathbb{R}^{B \times H' \times W' \times C}, \quad 
H' = \frac{H}{p}, \ \ W' = \frac{W}{p}. \label{eq:token_shape}
\end{align}
Here, $p$ denotes the patch size and $C$ is the embedding dimension. We apply a shared and frozen absolute positional encoding $\mathbf{P}_{\mathrm{abs}}$ to both streams. To maintain consistent spatial priors under varying input resolutions, $\mathbf{P}_{\mathrm{abs}}$ is resized via bicubic interpolation and then added to the token embeddings.

The feature extraction stage consists of $L$ stacked Transformer blocks. Following the multi-level feature tapping configuration, we divide the encoder into $S$ stages according to the selected tap indices $\{t_s\}_{s=1}^{S}$, where $S=4$ by default. The $s$-th stage is defined as the sequence of Transformer blocks from block $t_{s-1}+1$ to block $t_s$, with $t_0=0$. Within each stage, the two modalities are structurally aligned to share the same unimodal modeling capacity and optimization path, while cross-modal interaction is introduced only through a small set of trainable PEFT modules. Given the input tokens of the $l$-th block, $\mathbf{X}_{l-1}, \mathbf{Y}_{l-1} \in \mathbb{R}^{B \times H' \times W' \times C}$, the shared Pre-LN unimodal modeling is formulated as

\begingroup\footnotesize
\begin{align}
\tilde{\mathbf{X}}_{l} &= \mathbf{X}_{l-1} + \mathrm{SA}\!\left(\mathrm{LN}(\mathbf{X}_{l-1})\right), \quad
\tilde{\mathbf{Y}}_{l} = \mathbf{Y}_{l-1} + \mathrm{SA}\!\left(\mathrm{LN}(\mathbf{Y}_{l-1})\right), \label{eq:preln_sa}\\
\bar{\mathbf{X}}_{l} &= \tilde{\mathbf{X}}_{l} + \mathrm{MLP}\!\left(\mathrm{LN}(\tilde{\mathbf{X}}_{l})\right), \quad
\bar{\mathbf{Y}}_{l} = \tilde{\mathbf{Y}}_{l} + \mathrm{MLP}\!\left(\mathrm{LN}(\tilde{\mathbf{Y}}_{l})\right). \label{eq:preln_mlp}
\end{align}
\endgroup

The parameters of $\mathrm{SA}(\cdot)$ and $\mathrm{MLP}(\cdot)$ are shared across the two streams (shared weights) and are frozen together with the backbone.

Rather than inserting trainable modules into every Transformer block, we adopt a stage-wise insertion strategy. For each selected stage, we place CPIA before the stage encoder and DGFM after the stage encoder:
\begin{itemize}
\item \textbf{CPIA (implicit semantic modulation):} before the input tokens enter the $s$-th stage, CPIA generates shared semantic prompts from the bimodal tokens and injects them into both streams via prompt-injected adapters, enabling cross-modal semantic modulation under a frozen backbone.
\item \textbf{DGFM (explicit stage-level fusion):} after the $s$-th stage produces its output features, DGFM adaptively fuses the paired modality features by leveraging their discrepancy cue to guide channel-wise and spatially adaptive fusion, yielding a compact multimodal representation for decoding.
\end{itemize}

Denote by $\mathbf{X}^{s}$ and $\mathbf{Y}^{s}$ the RGB and auxiliary features output from the $s$-th selected stage, respectively. DGFM produces the fused feature $\mathbf{F}^{s}$ for each stage. The multi-level fused features $\{\mathbf{F}^{s}\}_{s=1}^{S}$ are then fed into the decoder head to produce the main prediction $\mathbf{P}$. During training, we additionally attach two lightweight modality-specific auxiliary heads to generate $P_{\mathrm{rgb}}$ and $P_{\mathrm{aux}}$ for MCRM supervision, while only the main fused branch is retained during inference. Ultimately, only the PEFT parameters $\Theta_{\mathrm{peft}}$ are updated, while $\Theta_{\text{backbone}}$ remains frozen throughout training.

\subsection{Cross-modal Prompt-Injected Adapter (CPIA)}\label{sec:cpia}
Under the frozen-backbone setting, relying solely on independent Transformer computations in the two streams leads to insufficient deep cross-modal semantic interaction. To address this issue, we propose the CPIA, which is inserted before each selected stage encoder and operates on the paired modality tokens. It introduces a lightweight mechanism based on \emph{prompt generation} and \emph{prompt-injected adapters}, enabling implicit semantic complementarity across modalities without disrupting the pretrained feature space. Specifically, CPIA consists of two components: a Cross-modal Prompt Generator and a Prompt-Injected Adapter.

\paragraph{Cross-modal Prompt Generator}
Given the paired modality tokens before entering the $s$-th stage, $\mathbf{X}^{s}_{\mathrm{in}}$ and $\mathbf{Y}^{s}_{\mathrm{in}} \in \mathbb{R}^{B \times H' \times W' \times C}$, CPG first projects each modality to a reduced dimension $C_p=\lfloor r_p C \rfloor$. The two projected features are then concatenated along the channel dimension, fused, and finally projected back to $C$ to obtain a shared semantic base $\mathbf{Z}^{s}$:
\begin{align}
\mathbf{X}^{\downarrow,s} &= \mathbf{X}^{s}_{\mathrm{in}}\mathbf{W}^{\mathrm{rgb}}_{d}, \qquad
\mathbf{Y}^{\downarrow,s} = \mathbf{Y}^{s}_{\mathrm{in}}\mathbf{W}^{\mathrm{aux}}_{d}, \label{eq:cpg_down} \\
\mathbf{Z}^{s} &= \mathrm{Up}\!\left(\mathrm{Fuse}\!\left([\mathbf{X}^{\downarrow,s}; \mathbf{Y}^{\downarrow,s}]\right)\right),
\label{eq:cpg_fuse}
\end{align}

Here, $\mathrm{Fuse}(\cdot)$ and $\mathrm{Up}(\cdot)$ denote linear mappings.
To mitigate the distribution shift between different modalities, we introduce a Task Feature Transformation (TFT) module that generates modality-specific prompts via channel-wise affine modulation. Specifically, we define
\begin{align}
\mathbf{P}^{s}_{\mathrm{rgb}} &= \mathbf{Z}^{s} + \left(\mathbf{Z}^{s} \odot \boldsymbol{\gamma}_{\mathrm{rgb}} + \boldsymbol{\beta}_{\mathrm{rgb}}\right), \\
\mathbf{P}^{s}_{\mathrm{aux}} &= \mathbf{Z}^{s} + \left(\mathbf{Z}^{s} \odot \boldsymbol{\gamma}_{\mathrm{aux}} + \boldsymbol{\beta}_{\mathrm{aux}}\right), \label{eq:tft}
\end{align}
where $\boldsymbol{\gamma}_{m}, \boldsymbol{\beta}_{m} \in \mathbb{R}^{C}$ are learnable channel-wise parameters for modality $m \in \{\mathrm{rgb}, \mathrm{aux}\}$ and are broadcast along the spatial dimensions $H' \times W'$.

\paragraph{Prompt-Injected Adapter}
Conventional adapters are designed for unimodal features. In CPIA, we inject the prompt vectors generated by CPG as a dynamic bias into the bottleneck of an adapter. Specifically, we insert a prompt-injected adapter with a bottleneck structure into each stream, and use the prompt as an additive modulation term on the reduced intermediate representation. For modality $m \in \{\mathrm{rgb}, \mathrm{aux}\}$, the adapter mapping is defined as

\begingroup\footnotesize
\begin{equation}
\mathrm{Adapter}(\mathbf{X}, \mathbf{P})=\mathbf{X}+\mathbf{W}_{\mathrm{up}}
\Big(\sigma\big(\mathbf{W}_{\mathrm{down}}\mathbf{X}+\mathbf{W}_{\mathrm{prompt}}\mathbf{P}\big)\Big),
\label{eq:prompt_adapter}
\end{equation}
\endgroup

where $\mathbf{W}_{\mathrm{down}} \in \mathbb{R}^{C \times d}$, $\mathbf{W}_{\mathrm{up}} \in \mathbb{R}^{d \times C}$, and $\mathbf{W}_{\mathrm{prompt}} \in \mathbb{R}^{C \times d}$ are learnable matrices. The bottleneck dimension is $d=\lfloor r_a C \rfloor$, and $\sigma(\cdot)$ denotes the ReLU activation. We further apply dropout to improve generalization. The CPIA outputs are then given by
\begin{align}
\mathbf{X}^{s}_{\mathrm{cpia}} &= \mathrm{Adapter}\!\left(\mathbf{X}^{s}_{\mathrm{in}}, \mathbf{P}^{s}_{\mathrm{rgb}}\right), \\
\mathbf{Y}^{s}_{\mathrm{cpia}} &= \mathrm{Adapter}\!\left(\mathbf{Y}^{s}_{\mathrm{in}}, \mathbf{P}^{s}_{\mathrm{aux}}\right).
\label{eq:cpia_out}
\end{align}

The modulated tokens $\mathbf{X}^{s}_{\mathrm{cpia}}$ and $\mathbf{Y}^{s}_{\mathrm{cpia}}$ are then fed into the $s$-th stage encoder. This design symmetrically injects cross-modal prompts into the bottlenecks of both streams, thereby establishing a stable semantic communication channel under the frozen-backbone constraint.

\begin{figure}[t]
\centering
\includegraphics[width=0.70\linewidth]{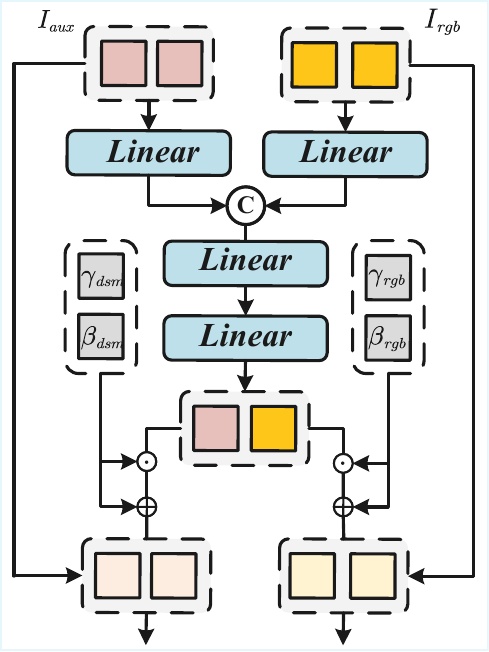}
\caption{Structure of the CPIA module. A shared semantic base is generated from paired RGB and auxiliary tokens, transformed into modality-specific prompts by TFT, and injected into lightweight bottleneck adapters for symmetric cross-modal semantic modulation before each selected stage.}
\label{fig:cpia}
\end{figure}

\begin{figure}[t]
\centering
\includegraphics[width=1.05\linewidth]{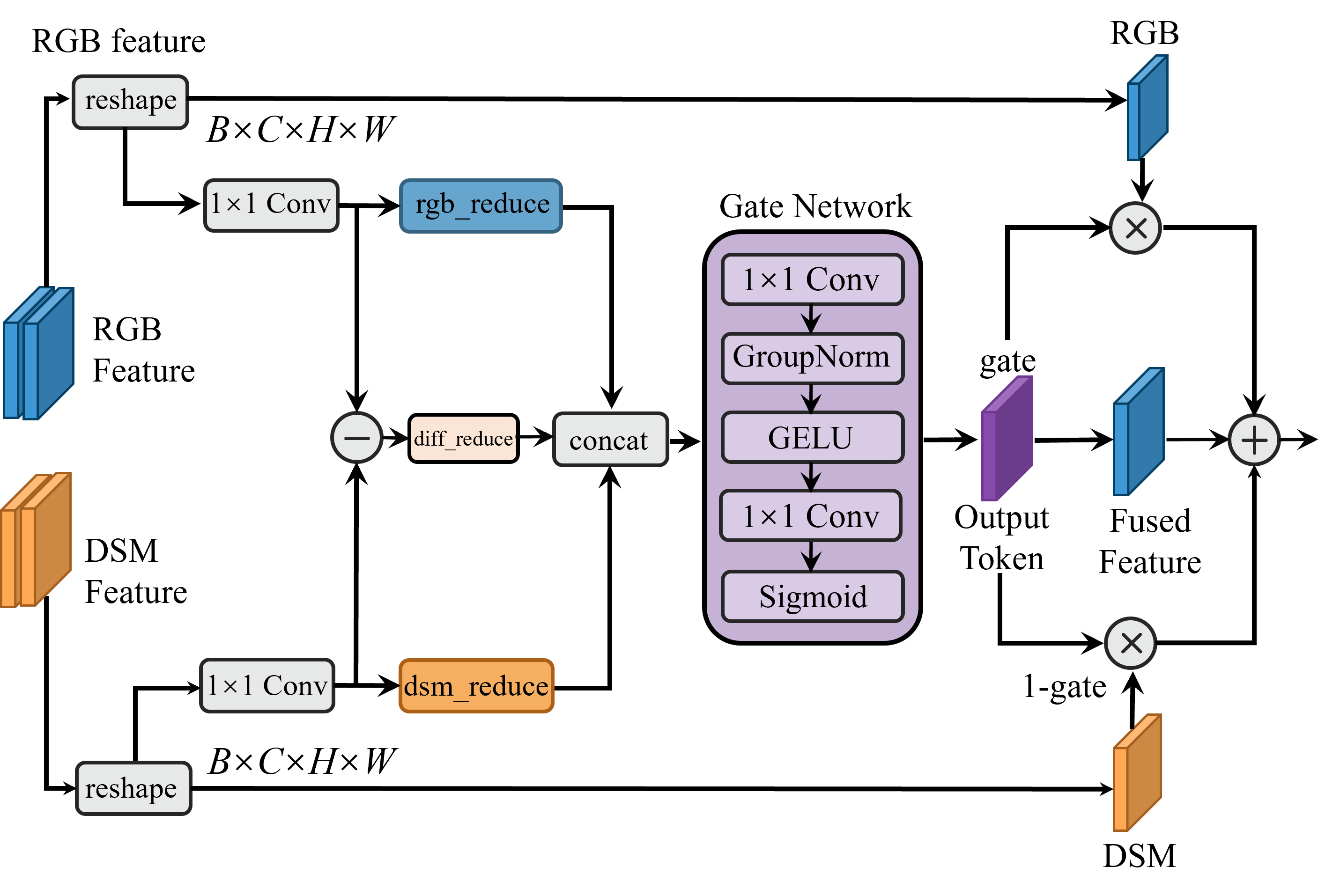}
\caption{Structure of the DGFM module. Reduced RGB and DSM features, together with their discrepancy cue, are fed into a lightweight gate network to predict an adaptive gating map, which dynamically fuses the two modalities into a compact multimodal representation.}
\label{fig:dgfm}
\end{figure}

\subsection{Difference-Guided Gated Fusion Module (DGFM)}\label{sec:DGFM}
To obtain compact and discriminative multimodal representations for decoding, we introduce a lightweight \emph{Difference-Guided Gated Fusion Module} (DGFM). For each selected stage, DGFM is applied after the stage encoder to fuse the paired optical and auxiliary features. Unlike more complicated fusion designs that introduce dedicated refinement branches, DGFM performs adaptive fusion in a simple yet effective manner by exploiting modality-specific responses together with their explicit discrepancy cue.

\paragraph{Compact Projection and Discrepancy Cue}
Given the paired features from the $s$-th selected encoder stage, $\mathbf{X}^{s}, \mathbf{Y}^{s} \in \mathbb{R}^{B \times L_s \times C_s}$ with $L_s = H_s W_s$, we first reshape them into 2-D feature maps, denoted by $\tilde{\mathbf{X}}^{s}, \tilde{\mathbf{Y}}^{s} \in \mathbb{R}^{B \times C_s \times H_s \times W_s}$. To reduce the computational cost of fusion, both modality features are projected into a compact channel space through $1 \times 1$ convolutions:
\begin{equation}
\mathbf{R}^{s}_{x} = \phi^{s}_{x}(\tilde{\mathbf{X}}^{s}), \qquad
\mathbf{R}^{s}_{y} = \phi^{s}_{y}(\tilde{\mathbf{Y}}^{s}),
\label{eq:DGFM_reduce}
\end{equation}
where $\phi^{s}_{x}(\cdot)$ and $\phi^{s}_{y}(\cdot)$ denote learnable channel reduction mappings. The reduced features satisfy $\mathbf{R}^{s}_{x}, \mathbf{R}^{s}_{y} \in \mathbb{R}^{B \times C'_s \times H_s \times W_s}$, where $C'_s < C_s$.

Based on the reduced features, we explicitly construct a discrepancy cue:
\begin{equation}
\mathbf{D}^{s} = \left|\mathbf{R}^{s}_{x} - \mathbf{R}^{s}_{y}\right|,
\label{eq:DGFM_diff}
\end{equation}
where $\mathbf{D}^{s} \in \mathbb{R}^{B \times C'_s \times H_s \times W_s}$ captures cross-modal disagreement and provides direct evidence for adaptive fusion.

\paragraph{Difference-Guided Adaptive Fusion}
The reduced modality features and their discrepancy cue are concatenated to predict a channel-wise and spatially adaptive gating map:
\begin{equation}
\mathbf{G}^{s} =
\sigma\!\left(
\phi^{s}_{g}\!\left([\mathbf{R}^{s}_{x}; \mathbf{R}^{s}_{y}; \mathbf{D}^{s}]\right)
\right),
\label{eq:DGFM_gate}
\end{equation}
where $[\cdot;\cdot]$ denotes channel-wise concatenation, $\sigma(\cdot)$ is the sigmoid function, and $\mathbf{G}^{s}\in\mathbb{R}^{B\times C_s\times H_s\times W_s}$ is the learned gating map. In practice, $\phi^{s}_{g}(\cdot)$ is implemented as a lightweight gating network consisting of a $1\times1$ convolution, GroupNorm, GELU, and a final $1\times1$ convolution. This design maps the concatenated compact features back to the original channel dimension, enabling channel-wise and spatially adaptive fusion over the full stage features.

The fused feature map is then formulated as
\begin{equation}
\tilde{\mathbf{F}}^{s}
=
\mathbf{G}^{s} \odot \tilde{\mathbf{X}}^{s}
+
(1-\mathbf{G}^{s}) \odot \tilde{\mathbf{Y}}^{s},
\label{eq:DGFM_fuse}
\end{equation}
where the gating map controls the contribution of the two modalities at each spatial location and channel. Finally, the fused feature map is reshaped back into token form:
\begin{equation}
\mathbf{F}^{s} = \mathrm{Reshape}(\tilde{\mathbf{F}}^{s}) \in \mathbb{R}^{B \times L_s \times C_s}.
\label{eq:DGFM_out}
\end{equation}

In this way, DGFM performs lightweight yet effective multimodal fusion by explicitly modeling cross-modal discrepancy and using it to guide adaptive feature selection. In the overall framework, CPIA facilitates pre-stage token-level semantic interaction under the frozen backbone, whereas DGFM performs post-stage feature-level adaptive fusion on the stage outputs.

\subsection{Modality-Conditional Random Masking (MCRM)}\label{sec:mcrm}
Although the proposed dual-stream architecture enables cross-modal interaction, the optimization process may still become dominated by the modality that provides stronger immediate predictive cues, thereby weakening the contribution of the complementary modality. To mitigate this issue, we introduce a training-only strategy termed \emph{Modality-Conditional Random Masking} (MCRM). The key idea is to simulate partial degradation in a modality-conditional manner: for a subset of training samples, only one modality is locally corrupted, while the counterpart remains intact. This design explicitly encourages the network to exploit complementary cross-modal evidence when one modality becomes unreliable.

\paragraph{Balanced Modality-Conditional Masking}
Given a mini-batch $\{(I_{\mathrm{rgb}}^{(i)}, I_{\mathrm{aux}}^{(i)})\}_{i=1}^{B}$, we first select $N=\lfloor rB \rfloor$ samples for masking, where $r\in[0,1]$ denotes the masking ratio. The selected samples are then partitioned into an RGB-masked subset and an Aux-masked subset in a nearly balanced manner:
\begin{align}
|B_{\mathrm{rgb}}| &= \left\lfloor \frac{N}{2} \right\rfloor, \notag\\
|B_{\mathrm{aux}}| &= N - |B_{\mathrm{rgb}}|, \notag\\
B_{\mathrm{full}} &= \{1,\ldots,B\}\setminus(B_{\mathrm{rgb}}\cup B_{\mathrm{aux}}). \tag{22}
\end{align}
We define a modality-conditional masking operator $\mathcal{M}(\cdot)$ that masks $K$ randomly sampled local rectangular regions on a single modality by setting the pixels within each region to zero. For each region, the area ratio, aspect ratio, and spatial location are randomly sampled from predefined ranges. The corrupted inputs are formulated as
\begin{equation}
(\tilde{I}_{\mathrm{rgb}}^{(i)}, \tilde{I}_{\mathrm{aux}}^{(i)})=
\begin{cases}
(\mathcal{M}(I_{\mathrm{rgb}}^{(i)}), I_{\mathrm{aux}}^{(i)}), & i\in B_{\mathrm{rgb}},\\
(I_{\mathrm{rgb}}^{(i)}, \mathcal{M}(I_{\mathrm{aux}}^{(i)})), & i\in B_{\mathrm{aux}},\\
(I_{\mathrm{rgb}}^{(i)}, I_{\mathrm{aux}}^{(i)}), & i\in B_{\mathrm{full}}.
\end{cases}
\tag{23}
\end{equation}
Therefore, for each selected sample, only one modality is locally degraded while the other remains intact. Such asymmetric corruption prevents the model from over-relying on a single dominant modality and promotes the learning of complementary multimodal representations under partial modality degradation.

\paragraph{Training Objective under MCRM}
Given the MCRM-corrupted input pair $(\tilde{I}_{\mathrm{rgb}}, \tilde{I}_{\mathrm{aux}})$, the network produces a fused prediction $P$ together with two modality-specific auxiliary predictions, denoted by $P_{\mathrm{rgb}}$ and $P_{\mathrm{aux}}$:
\begin{equation}
(P, P_{\mathrm{rgb}}, P_{\mathrm{aux}})=F(\tilde{I}_{\mathrm{rgb}}, \tilde{I}_{\mathrm{aux}}). \tag{24}
\end{equation}
The fused prediction is supervised by the standard cross-entropy loss:
\begin{equation}
\mathcal{L}_{\mathrm{main}}=\mathcal{L}_{\mathrm{ce}}(P, Y_{\mathrm{gt}}). \tag{25}
\end{equation}

To explicitly guide the modality-specific branches to compensate for the corrupted modality, we further impose auxiliary supervision on pixels that remain misclassified by the fused branch. Specifically, we define a hard-pixel set based on the fused prediction as
\begin{equation}
\Omega=\{u \mid \arg\max_c P(u,c)\neq Y_{\mathrm{gt}}(u)\}. \tag{26}
\end{equation}
Only pixels in $\Omega$ are used to supervise the auxiliary branches, while correctly predicted pixels are ignored. The auxiliary loss is thus defined as
\begin{equation}
\mathcal{L}_{\mathrm{aux}}=
\mathcal{L}_{\mathrm{ce}}(P_{\mathrm{rgb}}, Y_{\mathrm{gt}};\Omega)+
\mathcal{L}_{\mathrm{ce}}(P_{\mathrm{aux}}, Y_{\mathrm{gt}};\Omega). \tag{27}
\end{equation}
This hard-pixel-guided design avoids redundant supervision on already well-recognized regions and instead encourages each modality-specific branch to provide complementary cues precisely where multimodal fusion remains insufficient.

The overall training objective is formulated as
\begin{equation}
\mathcal{L}_{\mathrm{total}}=
\mathcal{L}_{\mathrm{main}}+\lambda_{\mathrm{aux}}\mathcal{L}_{\mathrm{aux}}, \tag{28}
\end{equation}
where $\lambda_{\mathrm{aux}}$ is the balancing coefficient for the auxiliary supervision. In this way, the fused branch is optimized to learn robust multimodal representations, while the modality-specific auxiliary branches are explicitly driven to compensate for locally corrupted inputs on error-prone pixels. Since MCRM is used only during training and introduces no extra inference overhead. The increased parameters in the full model mainly come from the lightweight auxiliary heads for MCRM supervision, which are removed at test time.

\begin{figure*}[t]
\centering
\includegraphics[width=\textwidth]{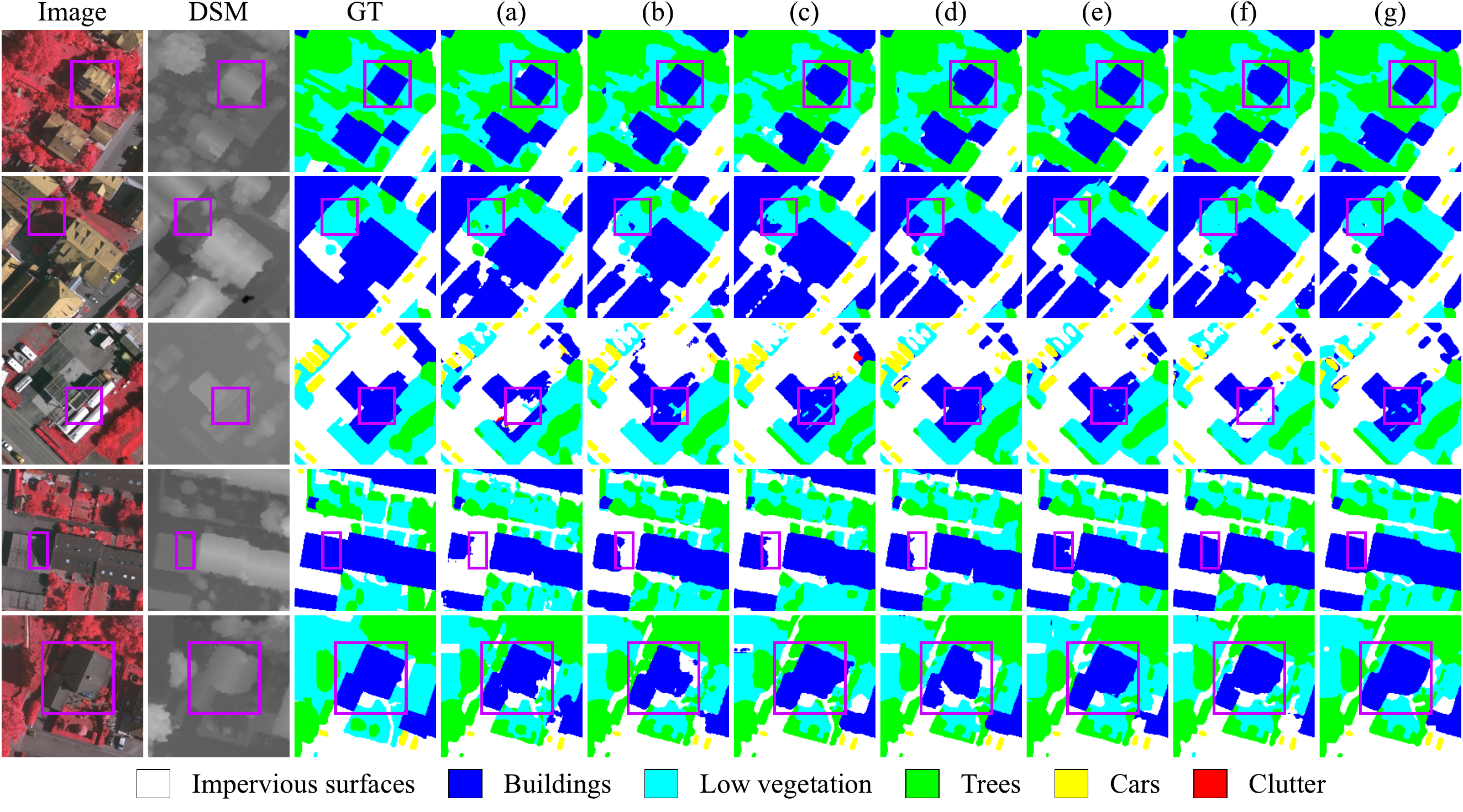}
\caption{Visualized comparisons on the Vaihingen test set with the size of 512 × 512. (a) U-Net, (b) ABCNet, (c) DC-Swin, (d) UNetFormer, (e) FTransUNet, (f) MANet, and (g) proposed MoBaNet. To highlight the differences, some purple boxes are added to all subfigures.}
\label{fig:vaihingen_compare_patch512}
\end{figure*}

\begin{figure*}[t]
\centering
\includegraphics[width=\textwidth]{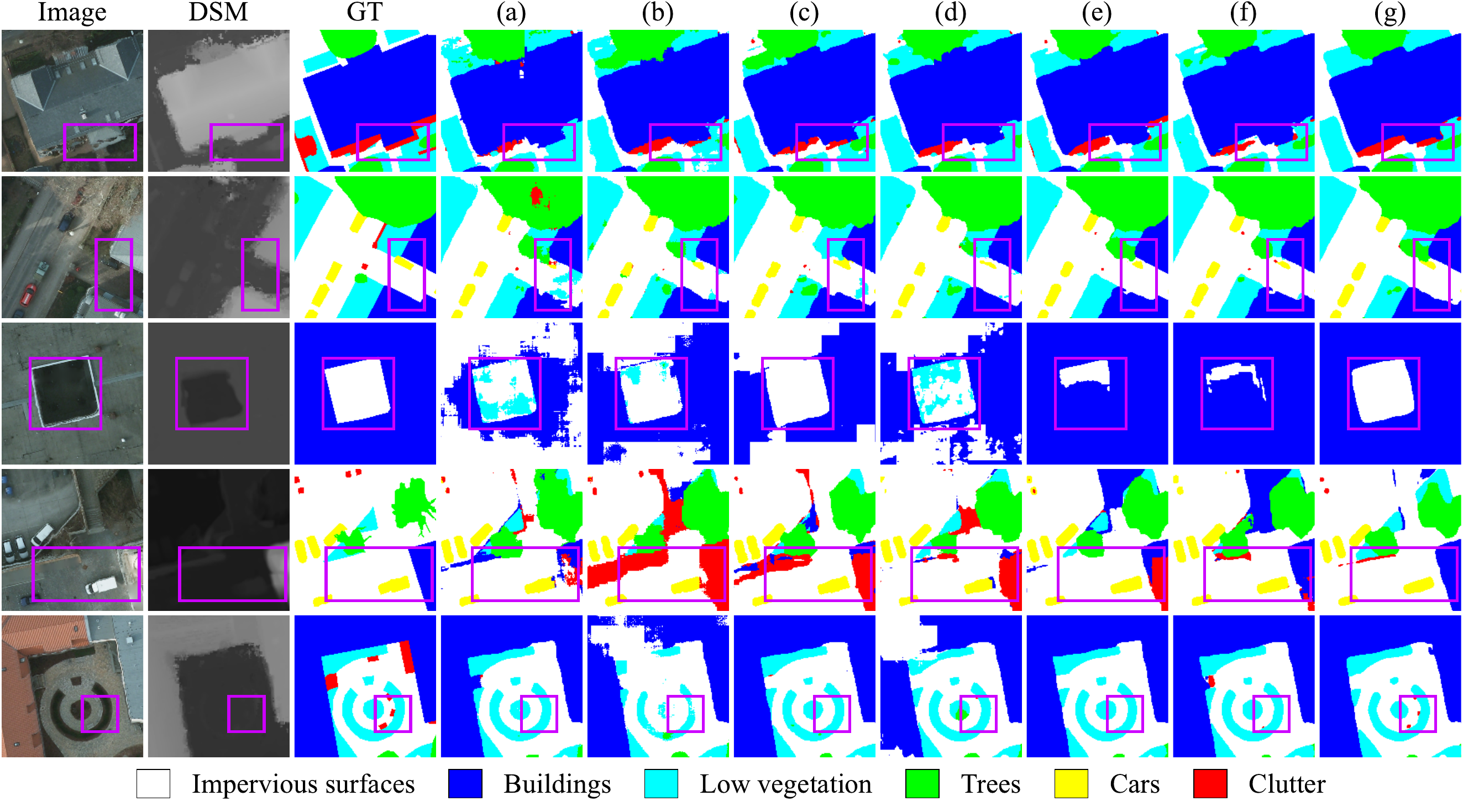}
\caption{Visualized comparisons on the Potsdam test set with the size of 512 × 512. (a) U-Net, (b) ABCNet, (c) DC-Swin, (d) UNetFormer, (e) FTransUNet, (f) MANet, and (g) proposed MoBaNet. To highlight the differences, some purple boxes are added to all subfigures.}
\label{fig:potsdam_compare_patch512}
\end{figure*}

\begin{table*}[t]
\centering
\caption{Quantitative comparison on the Vaihingen dataset. Best results are shown in bold, and second-best results are underlined(\%).}
\label{tab:perf_comp_vaihingen}
\setlength{\tabcolsep}{6pt}
\renewcommand{\arraystretch}{1.05}
\begin{tabular}{l l cccccc cc}
\toprule
Method & Backbone & Imp. & Bui. & Low. & Tre. & Car & OA & mF1 (\%) & mIoU (\%) \\
\midrule
U-Net~\cite{navab_u-net_2015}          & --         & 86.63 & 90.56 & 76.84 & 83.75 & 72.80 & 80.73 & 82.12 & 70.17 \\
SegNet~\cite{badrinarayanan_segnet_2017}          & --         & 86.72 & 91.31 & 78.82 & 84.87 & 69.66 & 81.55 & 82.27 & 70.55 \\
FuseNet~\cite{lai_fusenet_2017}          & VGG16         & 91.66 & 96.28 & 78.98 & 90.28 & 81.37 & 90.51 & 87.71 & 78.71 \\
PSPNet~\cite{zhao_pyramid_2017}          & ResNet101         & 92.79 & 95.46 & 84.51 & 89.94 & 88.61 & 90.85 & 90.26 & 82.58 \\
ABCNet~\cite{li_abcnet_2021}           & ResNet18         & \underline{96.67} & 95.44  & 83.67 & 89.61 & 87.19  & 91.14 & 90.51 & 83.04 \\
DC-Swin~\cite{wang_novel_2022}          & Swin-B      & \textbf{96.68} & 95.85 & 84.38 & 89.59 & 86.31 & 91.33 & 90.56 & 83.13 \\
MAResU-Net~\cite{li_multistage_2022}      & ResNet18      & 91.97 & 95.04 & 83.74 & 89.35 & 78.28 & 90.05 & 87.68 & 80.75 \\
UNetFormer~\cite{wang_unetformer_2022}       & ResNet18     & 96.57 & 95.20 & 83.43 & 89.35 & 86.15 & 90.76 & 90.14 & 82.44 \\
\midrule
vFuseNet~\cite{audebert_beyond_2018}         & VGG16         & 91.0 & 94.4 & 84.5 & 89.9 & 86.3 & 90.0 & 89.22 & -- \\
CMFNet~\cite{ma_crossmodal_2022}          & VGG16      & 92.36 & \textbf{97.17} & 80.37 & 90.82 & 85.47 & 91.40 & 89.48 & 81.44 \\
CMGFNet~\cite{hosseinpour_cmgfnet_2022}          & ResNet34      & 92.81 & 96.20 & 84.08 & 90.26 & 85.02 & 90.98 & 89.67 & 81.60 \\
TMFNet~\cite{liu_transformer-based_2024}           & ResNet34      & 91.25 & 95.70 & 80.87 & 88.82 & 84.60 & 89.71 & 88.25 & 78.70 \\
FTransUNet~\cite{ma_multilevel_2024}       & R50-ViT-B     & 92.89 & 96.02 & 84.70 & 89.80 & 86.94 & 91.05 & 90.07 & 82.18 \\
MANet~\cite{ma_unified_2025}    & ViT-B   & 92.79 & 96.14 & 84.15 & 90.57 & 86.78 & 91.35 & 90.33 & 82.60 \\
\midrule
\multirow{2}{*}{MoBaNet (SAM)}
 & ViT-B & 93.21 & 96.51 & \textbf{85.72} & 91.58 & \textbf{90.06} & 91.65 & 90.97 & 83.66 \\
 & ViT-L & 94.25 & 96.35 & 84.64 & \textbf{92.07} & 89.14 & \underline{91.80} & \underline{91.29} & \underline{84.17} \\
\midrule
\multirow{2}{*}{MoBaNet (DINOv2)}
 & ViT-B & 93.40 & 96.78 & \underline{85.54} & 90.55 & \underline{89.52} & 91.78 & 91.16 & 83.97 \\
 & ViT-L & 94.18 & \underline{96.90} & 84.58 & \underline{92.03} & 88.32 & \textbf{91.90} & \textbf{91.39} & \textbf{84.37} \\
\bottomrule
\end{tabular}
\end{table*}

\begin{table*}[t]
\centering
\caption{Quantitative comparison on the Potsdam dataset. Best results are shown in bold, and second-best results are underlined(\%).}
\label{tab:perf_comp_potsdam}
\setlength{\tabcolsep}{6pt}
\renewcommand{\arraystretch}{1.05}
\begin{tabular}{l l cccccc cc}
\toprule
Method & Backbone & Imp. & Bui. & Low. & Tre. & Car & OA & mF1 (\%) & mIoU (\%) \\
\midrule
U-Net~\cite{navab_u-net_2015}          & --         & 86.62 & 91.96 & 81.52 & 84.09 & 83.55 & 81.93 & 85.55 & 74.92 \\
SegNet~\cite{badrinarayanan_segnet_2017}          & --         & 86.77 & 92.43 & 79.59 & 81.79 & 82.97 & 81.14 & 84.71 & 73.75 \\
FuseNet~\cite{lai_fusenet_2017}          & VGG16         & 92.64 & 97.48 & 87.31 & 85.14 & 96.10 & 90.58 & 91.60 & 84.86 \\
PSPNet~\cite{zhao_pyramid_2017}          & ResNet101         & 93.36 & 96.97 & 87.75 & 88.50 & 95.42 & 91.08 & 92.40 & 84.88 \\
ABCNet~\cite{li_abcnet_2021}           & ResNet18         & 92.26 & 95.13 & 85.82 & 87.96 & 94.80 & 89.70 & 91.19 & 84.03 \\
DC-Swin~\cite{wang_novel_2022}          & Swin-B      & 93.65 & 96.31 & 87.16 & 88.31 & 95.27 & 90.86 & 92.14 & 85.64 \\
MAResU-Net~\cite{li_multistage_2022}      & ResNet18      & 91.4 & 95.6 & 85.8 & 86.6 & 93.3 & 89.0 & 90.5 & 83.9 \\
UNetFormer~\cite{wang_unetformer_2022}       & ResNet18     & 92.80 & 95.22 & 86.83 & 88.19 & 95.88 & 90.28 & 91.78 & 85.03 \\
\midrule
vFuseNet~\cite{audebert_beyond_2018}         & VGG16         & 92.7 & 96.3 & 87.3 & 88.5 & 95.4 & 90.6 & 92.0 & -- \\
CMFNet~\cite{ma_crossmodal_2022}          & VGG16      & 92.84 & \textbf{97.63} & 88.00 & 87.40 & 95.68 & 91.16 & 92.10 & 85.63 \\
CMGFNet~\cite{hosseinpour_cmgfnet_2022}          & ResNet34      & 92.62 & 97.20 & 86.46 & 88.55 & 94.46 & 90.59 & 91.86 & 85.18 \\
TMFNet~\cite{liu_transformer-based_2024}           & ResNet34      & 92.83 & 96.20 & 89.86 & 89.09 & 91.80 & 91.05 & 92.21 & 85.71 \\
FTransUNet~\cite{ma_multilevel_2024}       & R50-ViT-B     & 92.70 & 97.24 & 87.53 & 88.64 & 94.39 & 90.78 & 92.10 & 85.56 \\
MANet~\cite{ma_unified_2025}    & ViT-B   & 93.94 & 97.46 & 89.72 & 86.41 & 95.17 & 90.50 & 91.82 & 85.08 \\
\midrule
\multirow{2}{*}{MoBaNet (SAM)}
 & ViT-B & \underline{95.15} & 96.64 & \textbf{90.84} & 87.45 & 94.83 & 91.27 & 92.58 & 86.38 \\
 & ViT-L & \textbf{95.64} & 97.12 & \underline{90.78} & 85.95 & 94.88 & 91.39 & 92.73 & 86.65 \\
\midrule
\multirow{2}{*}{MoBaNet (DINOv2)}
 & ViT-B & 94.06 & 97.57 & 88.78 & \textbf{90.29} & \underline{96.29} & \underline{92.11} & \underline{93.40} & \underline{87.80} \\
 & ViT-L & 94.16 & \underline{97.58} & 89.08 & \underline{90.17} & \textbf{96.66} & \textbf{92.23} & \textbf{93.53} & \textbf{88.03} \\
\bottomrule
\end{tabular}
\end{table*}

\section{Experiments}\label{sec:exp}
\subsection{Datasets}
We evaluate our method on two multimodal benchmarks: ISPRS Vaihingen and ISPRS Potsdam. Both datasets comprise high-resolution true orthophotos (TOP) and digital surface models (DSM), with pixel-wise annotations for six classes: impervious surfaces, building, low vegetation, tree, car, and clutter.

\paragraph{ISPRS Vaihingen}
The ISPRS Vaihingen dataset was acquired over the town of Vaihingen, Germany, covering a typical small urban area dominated by detached houses and multi-story residential buildings. It contains 33 TOP tiles with varying spatial extents at a ground sampling distance (GSD) of approximately 9 cm. The optical imagery includes three spectral channels: near-infrared (NIR), red (R), and green (G), and DSM is provided as the auxiliary modality. Following the official ISPRS split, we use 16 tiles for training and the remaining 17 tiles for testing.

\paragraph{ISPRS Potsdam}
The ISPRS Potsdam dataset depicts a typical historical urban scene with dense building blocks, narrow streets, and complex urban textures. It has a higher spatial resolution with a GSD of approximately 5 cm and consists of 38 TOP tiles of a uniform size ($6000 \times 6000$ pixels). The optical imagery contains four spectral channels (NIR, R, G, and B), and DSM is provided accordingly. Following the official split protocol, we use 24 tiles for training and the remaining 14 tiles for testing and evaluation.

\subsection{Implementation Details}
All experiments are implemented in PyTorch and conducted on a single NVIDIA GeForce RTX 4090 GPU (24GB memory). We employ pretrained DINOv2 and SAM as the feature extraction backbones for comparison. The models are optimized using AdamW with an initial learning rate of $3\times10^{-4}$, a weight decay of 0.01, and a batch size of 12 (reduced to 6 when using ViT-L due to its higher memory footprint). The learning rate schedule consists of a linear warm-up for the first 5 epochs followed by cosine annealing. During training, we apply random cropping with a crop size of $256\times256$, together with random flipping and mirroring for data augmentation. All models are trained for 50 epochs, with 1000 iterations per epoch. We adopt UperNet~\cite{xiao_unified_2018} as the segmentation decoder to produce pixel-wise prediction masks. For the auxiliary modality, DSM is independently min--max normalized to $[0,1]$ on each tile in both training and testing. For RGB images, pixel values are first scaled to $[0,1]$; when the encoder is from the DINOv2 family, we further apply ImageNet normalization, otherwise no additional ImageNet standardization is performed. For reproducibility, the random seed is fixed to 42.

We evaluate segmentation performance using Overall Accuracy (OA), mean F1-score (mF1), and mean Intersection-over-Union (mIoU). Following the standard evaluation protocol on ISPRS benchmarks, OA is computed as pixel-wise accuracy over all classes, including the background category, whereas mF1 and mIoU are averaged over the five foreground classes only. This setting enables a fair comparison with existing state-of-the-art approaches.

\subsection{Performance Comparison}
To comprehensively evaluate the effectiveness and generality of the proposed MoBaNet for multimodal remote sensing semantic segmentation, we benchmark it against representative baselines spanning different modeling paradigms, including classical CNN-based segmentation networks, strong Transformer-based segmenters, and multimodal frameworks equipped with explicit cross-modal fusion mechanisms. Specifically, UNet~\cite{navab_u-net_2015}, SegNet~\cite{badrinarayanan_segnet_2017} FuseNet~\cite{lai_fusenet_2017}, PSPNet\cite{zhao_pyramid_2017}, ABCNet~\cite{li_abcnet_2021}, DC-Swin~\cite{wang_novel_2022}, MAResU-Net~\cite{li_multistage_2022}, and UNetFormer~\cite{wang_unetformer_2022} are adopted as strong optical-only baselines. These methods take only optical imagery as input, characterizing the attainable performance without geometric cues and serving as a reference for quantifying the gains brought by multimodal fusion. In contrast, vFuseNet~\cite{audebert_beyond_2018}, CMFNet~\cite{ma_crossmodal_2022}, CMGFNet~\cite{hosseinpour_cmgfnet_2022}, TMFNet~\cite{liu_transformer-based_2024}, FTransUNet~\cite{ma_multilevel_2024}, and MANet~\cite{ma_unified_2025} further incorporate DSM on top of optical inputs and perform explicit fusion, representing typical designs for multimodal segmentation. To promote a fair comparison, we adopt consistent DSM co-registration/alignment, normalization, preprocessing, dataset splits, and evaluation metrics whenever possible, thereby reducing potential biases caused by inconsistent modality handling. Moreover, to examine the backbone-agnostic property of our framework, we implement and evaluate MoBaNet with SAM and DINOv2, demonstrating its stability and generalization across different backbones.
Overall, the comparative quantitative results are reported in Tables~\ref{tab:perf_comp_vaihingen} and~\ref{tab:perf_comp_potsdam}.

\paragraph{Performance Comparison on the Vaihingen Dataset} Table~\ref{tab:perf_comp_vaihingen} summarizes the quantitative results on the ISPRS Vaihingen test set. MoBaNet achieves the best overall performance in terms of OA, mF1, and mIoU, demonstrating that a parameter-efficient and modality-balanced symmetric fusion scheme can effectively exploit the complementarity between optical imagery and DSM while keeping the VFM backbone largely frozen. Specifically, MoBaNet (DINOv2-ViT-L) attains 91.90\% OA, 91.39\% mF1, and 84.37\% mIoU, which are the best results in Table~\ref{tab:perf_comp_vaihingen}. Compared with the strong optical-only baseline DC-Swin, MoBaNet improves mIoU by 1.24 percentage points, indicating that DSM-provided geometric cues help reduce appearance-induced inter-class confusion and boundary deviations. Furthermore, MoBaNet outperforms the recent multimodal method MANet (ViT-B) by 1.77 percentage points in mIoU (84.37\% vs. 82.60\%), suggesting that under a constrained trainable-parameter budget, the proposed interaction-and-fusion design (CPIA/DGFM) together with the bias-mitigating training strategy (MCRM) can better activate discriminative cues from the auxiliary modality instead of converging to RGB-dominant shortcut solutions.   

Fig.~\ref{fig:vaihingen_compare_patch512} shows qualitative comparisons on the Vaihingen test set. MoBaNet (Fig.~\ref{fig:vaihingen_compare_patch512}(g)) produces predictions that better preserve object integrity and yield sharper boundary delineation. In shadowed regions or areas with pronounced illumination variation, optical-only methods are more susceptible to spectral perturbations, often resulting in local misclassification, boundary bleeding, or fragmented masks. By contrast, MoBaNet generates more coherent segments with fewer artifacts for challenging classes affected by occlusion and texture ambiguity (e.g., low vegetation and car), reflecting a more effective utilization of DSM structural and height cues through cross-modal interaction and discrepancy-guided adaptive fusion.

\paragraph{Performance Comparison on the Potsdam Dataset}
Table~\ref{tab:perf_comp_potsdam} reports the quantitative comparisons on the ISPRS Potsdam test set. MoBaNet again achieves the best performance across OA, mF1, and mIoU, indicating that the proposed framework remains effective in higher-resolution urban scenes with richer textures and more complex object layouts. In particular, MoBaNet (DINOv2-ViT-L) reaches 92.23\% OA, 93.53\% mF1, and 88.03\% mIoU, which are the best results in Table~\ref{tab:perf_comp_potsdam}. Compared with the strong optical-only baseline DC-Swin, MoBaNet improves mIoU by 2.39 percentage points (88.03\% vs. 85.64\%), suggesting that DSM-derived geometric/height cues are especially beneficial for reducing confusion among visually similar categories such as impervious surfaces, building, and low vegetation in dense urban areas. Moreover, MoBaNet surpasses the recent multimodal baseline MANet by 2.95 percentage points in mIoU (88.03\% vs. 85.08\%), implying that effective exploitation of auxiliary modalities requires not only a fusion architecture but also mechanisms that encourage consistent cross-modal contribution under parameter-efficient adaptation. Notably, MoBaNet achieves strong performance on challenging categories such as low vegetation and car (e.g., 89.08\% and 96.66\%, respectively), further reflecting its improved discriminative capability for fine-grained targets and ambiguous regions.

Fig.~\ref{fig:potsdam_compare_patch512} presents qualitative comparisons on the Potsdam test set. Owing to its higher spatial resolution and intricate urban textures, accurate segmentation requires preserving small objects, recovering regular geometric structures, and suppressing spurious predictions in cluttered backgrounds. As highlighted by the purple boxes, competing methods tend to miss or confuse small targets and produce noisy or broken masks in challenging regions, whereas MoBaNet yields cleaner predictions with more coherent regions and boundaries that better follow geometric contours. In addition, MoBaNet maintains clearer delineation at class transitions (e.g., road--building and vehicle--road interfaces), demonstrating improved robustness and fine-detail control in dense urban scenes.

\begin{table*}[t]
\centering
\caption{Computational complexity analysis measured using a $256\times256$ image on a single NVIDIA GeForce RTX 4090 GPU. For MoBaNet, the reported results are obtained using the DINOv2-ViT-B backbone. mIoU values are reported on the Potsdam dataset. A horizontal rule separates optical-only and multimodal methods. Bold values indicate the best results.}
\label{tab:efficiency_comparison}
\setlength{\tabcolsep}{6pt}
\renewcommand{\arraystretch}{1.05}
\begin{tabular}{l cccc c}
\toprule
Method & Params (M) & Latency (ms)$\downarrow$ & FPS$\uparrow$ & Memory (MB)$\downarrow$ & mIoU (\%)$\uparrow$ \\
\midrule
ABCNet      & 13.67  & \textbf{3.55}  & \textbf{281.61} & 194.59 & 84.03 \\
DC-Swin     & 118.39 & 14.59 & 68.56 & 603.83 & 85.64 \\
UNetFormer  & 11.72 & 3.58 & 279.51 & \textbf{186.92} & 85.03 \\
\midrule
CMGFNet     & 43.68 & 13.13 & 76.18 & 357.94 & 85.18 \\
FTransUNet  & 203.40 & 38.90 & 25.71 & 930.35 & 85.56 \\
MANet       & 20.42 & 105.36 & 9.49 & 1741.83 & 85.08 \\
MoBaNet     & \textbf{6.18} & 13.97 & 71.60 & 402.99 & \textbf{87.80} \\
\bottomrule
\end{tabular}
\end{table*}

\begin{table*}[t]
\centering
\caption{Ablation study of the proposed key designs on the ISPRS Potsdam dataset. 
``Base'' denotes the frozen VFM dual-stream baseline with a trainable auxiliary patch embedding and the same decoder head, without any of the proposed designs. 
``Params'' reports the number of trainable parameters (M). The best results are in bold and the second-best are underlined.}
\label{tab:ablation_components_potsdam}
\setlength{\tabcolsep}{6pt}
\renewcommand{\arraystretch}{1.05}
\begin{tabular}{l cccc ccc}
\toprule
Variant & CPIA & DGFM & MCRM & Params (M) & OA (\%) & mF1 (\%) & mIoU (\%) \\
\midrule
Base                                &  &  &  & 1.91 & 90.47 & 91.74 & 84.95 \\
Base + CPIA                         & \checkmark &  &  & 3.99 & 91.43 & 92.57 & 86.33 \\
Base + CPIA + DGFM                  & \checkmark & \checkmark &  & 5.00 & 91.73 & 93.02 & 87.15 \\
Base + CPIA + DGFM + MCRM (Full)    & \checkmark & \checkmark & \checkmark & 6.18 & \textbf{92.11} & \textbf{93.40} & \textbf{87.80} \\
\bottomrule
\end{tabular}
\end{table*}

\begin{table*}[t]
\centering
\caption{Modality-missing evaluation at inference time. ``RGB-only'' sets DSM to zeros. All models are trained with full modalities unless otherwise specified. ``Drop ($\downarrow$)'' is computed as the performance difference between the RGB+DSM and RGB-only settings.}
\label{tab:ablation_modality_missing}
\setlength{\tabcolsep}{6pt}
\renewcommand{\arraystretch}{1.05}
\begin{tabular}{l l ccc ccc}
\toprule
\multirow{2}{*}{Dataset} & \multirow{2}{*}{Input} 
& \multicolumn{3}{c}{MoBaNet (Full)} 
& \multicolumn{3}{c}{Reference Baseline} \\
\cmidrule(lr){3-5}\cmidrule(lr){6-8}
& & OA (\%) & mF1 (\%) & mIoU (\%) & OA (\%) & mF1 (\%) & mIoU (\%) \\
\midrule
\multirow{3}{*}{Potsdam} 
& RGB + DSM   & 92.11 & 93.40 & 87.80 & 91.79 & 93.09 & 87.26 \\
& RGB-only    & 91.36 & 92.74 & 86.63 & 72.38 & 76.97 & 64.20 \\
& Drop ($\downarrow$) & 0.75 & 0.66 & 1.17 & 19.41 & 16.12 & 23.06 \\

\bottomrule
\end{tabular}
\end{table*}

\subsection{Model Complexity Analysis}\label{sec:complexity}
To evaluate the computational complexity and practical efficiency of the proposed MoBaNet, we compare a representative MoBaNet configuration based on the DINOv2-ViT-B backbone with representative optical-only and multimodal baselines in terms of trainable parameters, inference latency, FPS, peak memory footprint, and mIoU, as reported in Table~\ref{tab:efficiency_comparison}. These metrics jointly reflect the trade-off between segmentation performance and deployment cost.

MoBaNet achieves the best segmentation accuracy among all compared methods, reaching 87.80\% mIoU, while requiring only 6.18M trainable parameters, which is also the smallest parameter count in the table. This result demonstrates that the proposed parameter-efficient adaptation strategy can effectively leverage the representation capability of the foundation backbone with minimal trainable overhead. Compared with representative multimodal baselines, MoBaNet is substantially more lightweight than CMGFNet (43.68M), FTransUNet (203.40M), and MANet (20.42M), while outperforming them in mIoU. In particular, MoBaNet exceeds FTransUNet and MANet by 2.24 and 2.72 percentage points, respectively.

From the perspective of deployment efficiency, MoBaNet also exhibits a favorable balance between performance and hardware cost. Its inference latency is 13.97 ms with a corresponding FPS of 71.60, which is close to that of the efficient multimodal baseline CMGFNet (13.13 ms, 76.18 FPS), while being markedly better than FTransUNet and MANet in both speed and memory usage. Although lightweight optical-only methods such as ABCNet and UNetFormer remain faster and more memory-efficient, they do not exploit complementary DSM information and therefore achieve lower segmentation accuracy than MoBaNet. Overall, these results indicate that MoBaNet provides a strong trade-off among segmentation accuracy, parameter efficiency, and inference cost, highlighting its practical potential for multimodal remote sensing segmentation under resource-constrained deployment settings.

\subsection{Ablation Study}\label{sec:ablation}
To evaluate the contribution of each proposed component and the robustness of the overall framework, we conduct two groups of ablation experiments under the same training protocol described in Sec.~\ref{sec:exp}, using DINOv2-ViT-B as the default backbone. Unless otherwise specified, all variants are evaluated in terms of OA, mF1, and mIoU, and the percentage of trainable parameters is additionally reported to reflect parameter efficiency.

\paragraph{Component Ablation}
We first investigate the contribution of each module by progressively introducing CPIA, DGFM, and MCRM on top of a frozen dual-stream baseline, denoted as \emph{Base}. Specifically, \emph{Base} consists of the frozen backbone with a trainable auxiliary patch embedding and the same decoder head, but without any of the proposed components. Starting from \emph{Base}, incorporating \textbf{CPIA} yields consistent improvements across all evaluation metrics, indicating that prompt-injected adapters effectively facilitate cross-modal semantic interaction under a largely frozen backbone. Further introducing \textbf{DGFM} brings additional gains, suggesting that discrepancy-guided adaptive fusion is beneficial for suppressing noisy responses and improving dense prediction quality. Finally, adding \textbf{MCRM} leads to the best overall performance, confirming that Modality-Conditional Random Masking helps mitigate RGB-dominant optimization and promotes a more balanced exploitation of complementary DSM cues. Overall, the full model achieves the best trade-off between segmentation accuracy and parameter efficiency, demonstrating the effectiveness of the proposed multimodal adaptation design.

\paragraph{Modality-Missing Robustness}
We further evaluate the robustness of the model under modality-missing inference to examine whether the learned representation is overly dependent on a single modality. The comparison is conducted on the Potsdam dataset. Specifically, two inference settings are considered: \textit{RGB+DSM} and \textit{RGB-only}, where the latter is obtained by setting DSM to zeros while keeping all network parameters unchanged. To isolate the effect of the proposed regularization, the reference model is the variant trained without MCRM, i.e., \emph{Base+CPIA+DGFM}. As shown in Table~\ref{tab:ablation_modality_missing}, the full model exhibits substantially smaller performance degradation when DSM is removed. In particular, under the \textit{RGB-only} setting, the performance drop of MoBaNet is 0.75\% in OA, 0.66\% in mF1, and 1.17\% in mIoU, whereas the reference baseline suffers much larger drops of 19.41\%, 16.12\%, and 23.06\%, respectively. These results indicate that MCRM effectively reduces the model's over-reliance on a single dominant modality and improves the robustness of multimodal representation learning under partial modality degradation. Notably, MCRM is applied only during training and introduces no additional computational overhead at inference time.

\begin{figure}[t]
\centering
\includegraphics[width=1\linewidth]{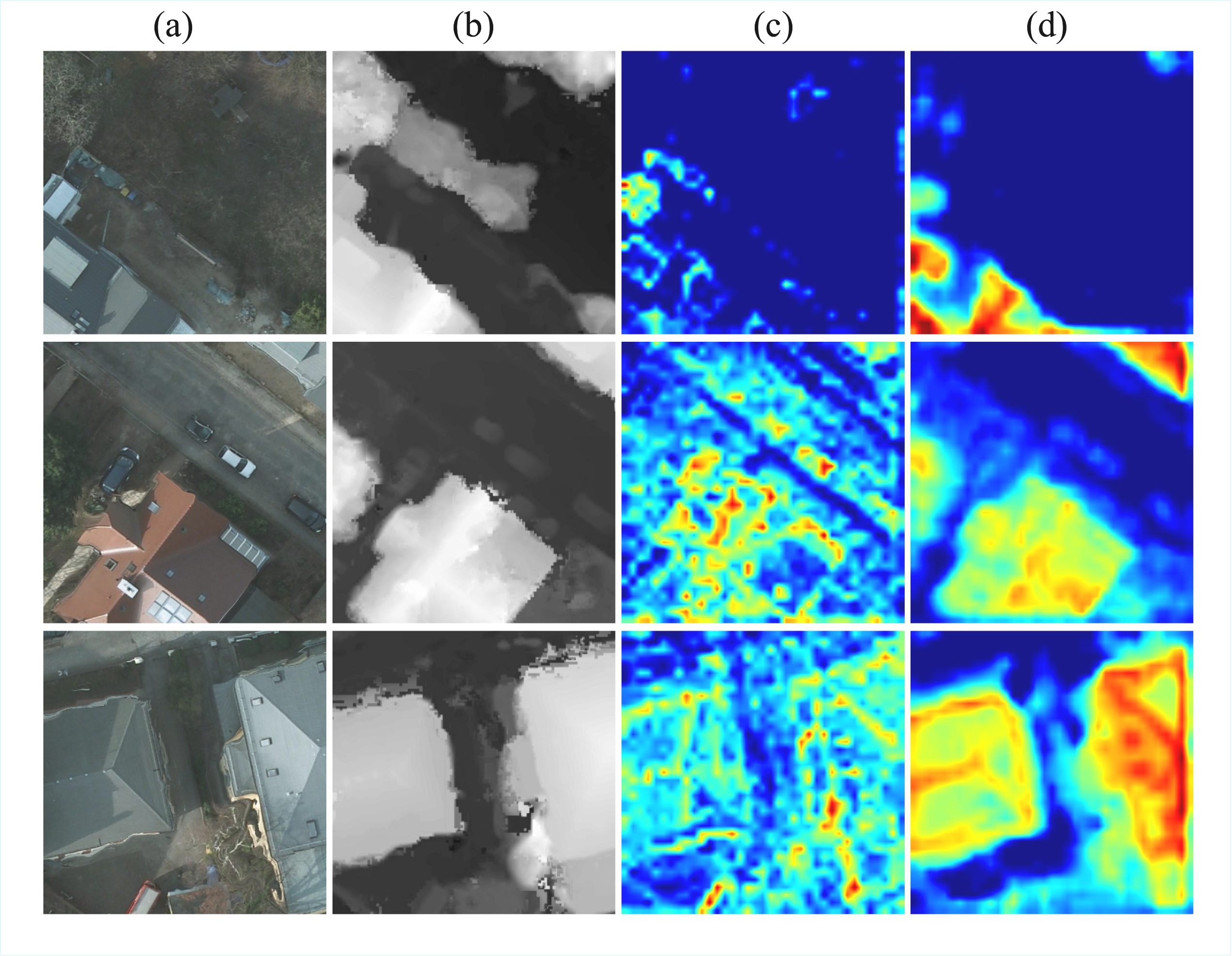}
\caption{Qualitative comparison of feature response visualizations on representative Potsdam samples for the \emph{Buildings} category. (a) RGB image. (b) DSM. (c) \emph{Base}. (d) Full model.}
\label{fig:heatmap}
\end{figure}

\paragraph{Qualitative Visualization}
To complement the quantitative analysis, Fig.~\ref{fig:heatmap} presents class-specific response maps for the \emph{Buildings} category on representative samples from the Potsdam dataset. Compared with \emph{Base}, the full model produces more compact, spatially coherent, and semantically concentrated activations over building regions, while suppressing irrelevant responses in the surrounding background. In particular, the highlighted regions generated by the full model align more closely with the spatial extent and boundaries of buildings under varying roof appearances and complex local contexts. These qualitative observations are consistent with the quantitative improvements reported in Table~\ref{tab:ablation_components_potsdam} and provide further visual evidence that the proposed design enhances multimodal representation quality.

\section{Discussion and Future Work}\label{sec:discussion}

This work presents MoBaNet, a unified parameter-efficient multimodal fine-tuning framework for remote sensing semantic segmentation. Built upon a largely frozen vision foundation model, MoBaNet integrates stage-wise cross-modal modulation, difference-guided fusion, and training-time robustness regularization, and demonstrates strong effectiveness on RGB--DSM segmentation benchmarks. As an early exploration of parameter-efficient multimodal adaptation for remote sensing with large vision backbones, this work also suggests several directions for further study.

\noindent\hangindent=2.2em\hangafter=1\hspace*{1.2em}1) \textit{Improving parameter-efficient adaptation modules:} This work adopts CPIA as a lightweight stage-wise prompt-injected adaptation mechanism to enable cross-modal interaction under a frozen backbone. Future research may explore more advanced or more efficient PEFT designs, such as stronger prompt-based, low-rank, or hybrid adaptation strategies, to further improve adaptation quality while reducing memory and optimization costs for large-scale multimodal models.

\noindent\hangindent=2.2em\hangafter=1\hspace*{1.2em}2) \textit{Improving multimodal fusion modules:} DGFM performs discrepancy-guided adaptive fusion to selectively integrate complementary cues from heterogeneous modalities at selected stages. The experimental results suggest that explicitly modeling cross-modal discrepancy is an effective way to suppress inconsistent responses and improve the quality of fused representations. Nevertheless, the current fusion design remains relatively lightweight, and its capacity for modeling more complex cross-modal relationships and higher-level complementary interactions may still be limited.

\noindent\hangindent=2.2em\hangafter=1\hspace*{1.2em}3) \textit{Addressing challenging categories and fine structures:} The experimental results show that MoBaNet improves segmentation quality in complex urban scenes, particularly in terms of region coherence and boundary delineation. However, challenging categories such as low vegetation, trees, and small objects like cars remain difficult, indicating that fine-grained discrimination and structural detail modeling are still challenging under complex backgrounds and class ambiguity.

\noindent\hangindent=2.2em\hangafter=1\hspace*{1.2em}4) \textit{Extending to broader remote sensing modalities and settings:} This study mainly focuses on RGB--DSM segmentation on two urban benchmarks. The effectiveness of the proposed framework on other multimodal remote sensing settings, such as multispectral, LiDAR, or SAR data, remains to be explored. In addition, more challenging scenarios, including cross-city transfer, cross-sensor generalization, and incomplete-modality conditions, would provide a broader testbed for evaluating the scalability and robustness of MoBaNet.

Overall, this work provides a concise yet effective foundation for parameter-efficient multimodal adaptation of vision foundation models in remote sensing. We expect that the proposed framework can be further extended to more diverse modality combinations, scene types, and dense prediction tasks in future research.

\section{Conclusion}\label{sec:conclusion}

In this paper, we proposed MoBaNet, a parameter-efficient modality-balanced symmetric fusion framework for multimodal remote sensing semantic segmentation. Built upon a largely frozen vision foundation model, MoBaNet combines a stage-wise dual-stream adaptation architecture with three key designs: CPIA for pre-stage cross-modal semantic modulation, DGFM for post-stage discrepancy-guided adaptive fusion, and MCRM for improving modality balance and robustness during training.

Extensive experiments on the ISPRS Vaihingen and Potsdam benchmarks demonstrated that MoBaNet achieves state-of-the-art segmentation performance while requiring only a small proportion of trainable parameters compared with full fine-tuning. The ablation and modality-missing analyses further verified that the proposed modules contribute complementary benefits in semantic interaction, reliable fusion, and robustness under partial modality degradation. These results suggest that MoBaNet provides an effective and efficient solution for multimodal adaptation of large vision backbones in remote sensing segmentation. In future work, we will further investigate its extension to more diverse modality combinations and broader cross-domain remote sensing scenarios.


%

\ifCLASSOPTIONcaptionsoff
  \newpage
\fi



\bibliographystyle{IEEEtran}
\bibliography{bibtex/bib/IEEEabrv,bibtex/bib/multimodalseg}

@article{niu_hybrid_2022,
	title = {Hybrid {Multiple} {Attention} {Network} for {Semantic} {Segmentation} in {Aerial} {Images}},
	volume = {60},
	issn = {1558-0644},
	url = {https://ieeexplore.ieee.org/document/9385072/},
	doi = {10.1109/TGRS.2021.3065112},
	urldate = {2026-02-15},
	journal = {IEEE Transactions on Geoscience and Remote Sensing},
	author = {Niu, Ruigang and Sun, Xian and Tian, Yu and Diao, Wenhui and Chen, Kaiqiang and Fu, Kun},
	year = {2022},
	pages = {1--18},
}

@inproceedings{zhao_pyramid_2017,
	address = {Honolulu, HI},
	title = {Pyramid {Scene} {Parsing} {Network}},
	isbn = {978-1-5386-0457-1},
	url = {http://ieeexplore.ieee.org/document/8100143/},
	doi = {10.1109/CVPR.2017.660},
	language = {en},
	urldate = {2026-02-15},
	booktitle = {2017 {IEEE} {Conference} on {Computer} {Vision} and {Pattern} {Recognition} ({CVPR})},
	publisher = {IEEE},
	author = {Zhao, Hengshuang and Shi, Jianping and Qi, Xiaojuan and Wang, Xiaogang and Jia, Jiaya},
	month = jul,
	year = {2017},
	pages = {6230--6239},
}

@article{badrinarayanan_segnet_2017,
	title = {{SegNet}: {A} {Deep} {Convolutional} {Encoder}-{Decoder} {Architecture} for {Image} {Segmentation}},
	volume = {39},
	issn = {1939-3539},
	shorttitle = {{SegNet}},
	url = {https://ieeexplore.ieee.org/document/7803544/},
	doi = {10.1109/TPAMI.2016.2644615},
	number = {12},
	urldate = {2026-02-13},
	journal = {IEEE Transactions on Pattern Analysis and Machine Intelligence},
	author = {Badrinarayanan, Vijay and Kendall, Alex and Cipolla, Roberto},
	month = dec,
	year = {2017},
	pages = {2481--2495},
}

@incollection{navab_u-net_2015,
	address = {Cham},
	title = {U-{Net}: {Convolutional} {Networks} for {Biomedical} {Image} {Segmentation}},
	volume = {9351},
	isbn = {978-3-319-24573-7 978-3-319-24574-4},
	shorttitle = {U-{Net}},
	url = {http://link.springer.com/10.1007/978-3-319-24574-4_28},
	doi = {10.1007/978-3-319-24574-4_28},
	language = {en},
	urldate = {2026-02-04},
	booktitle = {Medical {Image} {Computing} and {Computer}-{Assisted} {Intervention} – {MICCAI} 2015},
	publisher = {Springer International Publishing},
	author = {Ronneberger, Olaf and Fischer, Philipp and Brox, Thomas},
	editor = {Navab, Nassir and Hornegger, Joachim and Wells, William M. and Frangi, Alejandro F.},
	year = {2015},
	note = {Series Title: Lecture Notes in Computer Science},
	pages = {234--241},
}

@article{audebert_beyond_2018,
	title = {Beyond {RGB}: {Very} high resolution urban remote sensing with multimodal deep networks},
	volume = {140},
	issn = {09242716},
	shorttitle = {Beyond {RGB}},
	url = {https://linkinghub.elsevier.com/retrieve/pii/S0924271617301818},
	doi = {10.1016/j.isprsjprs.2017.11.011},
	language = {en},
	urldate = {2026-02-04},
	journal = {ISPRS Journal of Photogrammetry and Remote Sensing},
	author = {Audebert, Nicolas and Le Saux, Bertrand and Lefèvre, Sébastien},
	month = jun,
	year = {2018},
	pages = {20--32},
}

@article{wang_novel_2022,
	title = {A {Novel} {Transformer} {Based} {Semantic} {Segmentation} {Scheme} for {Fine}-{Resolution} {Remote} {Sensing} {Images}},
	volume = {19},
	issn = {1558-0571},
	url = {https://ieeexplore.ieee.org/document/9681903/},
	doi = {10.1109/LGRS.2022.3143368},
	urldate = {2026-02-04},
	journal = {IEEE Geoscience and Remote Sensing Letters},
	author = {Wang, Libo and Li, Rui and Duan, Chenxi and Zhang, Ce and Meng, Xiaoliang and Fang, Shenghui},
	year = {2022},
	pages = {1--5},
}

@article{li_multistage_2022,
	title = {Multistage {Attention} {ResU}-{Net} for {Semantic} {Segmentation} of {Fine}-{Resolution} {Remote} {Sensing} {Images}},
	volume = {19},
	issn = {1558-0571},
	url = {https://ieeexplore.ieee.org/document/9378788/},
	doi = {10.1109/LGRS.2021.3063381},
	urldate = {2026-02-04},
	journal = {IEEE Geoscience and Remote Sensing Letters},
	author = {Li, Rui and Zheng, Shunyi and Duan, Chenxi and Su, Jianlin and Zhang, Ce},
	year = {2022},
	pages = {1--5},
}

@article{gao_global_2024,
	title = {Global feature-based multimodal semantic segmentation},
	volume = {151},
	issn = {00313203},
	url = {https://linkinghub.elsevier.com/retrieve/pii/S0031320324000918},
	doi = {10.1016/j.patcog.2024.110340},
	language = {en},
	urldate = {2026-02-03},
	journal = {Pattern Recognition},
	author = {Gao, Suining and Yang, Xiubin and Jiang, Li and Fu, Zongqiang and Du, Jiamin},
	month = jul,
	year = {2024},
	pages = {110340},
}

@article{wang_imbalance_2023,
	title = {Imbalance knowledge-driven multi-modal network for land-cover semantic segmentation using aerial images and {LiDAR} point clouds},
	volume = {202},
	issn = {09242716},
	url = {https://linkinghub.elsevier.com/retrieve/pii/S092427162300179X},
	doi = {10.1016/j.isprsjprs.2023.06.014},
	language = {en},
	urldate = {2026-02-03},
	journal = {ISPRS Journal of Photogrammetry and Remote Sensing},
	author = {Wang, Yameng and Wan, Yi and Zhang, Yongjun and Zhang, Bin and Gao, Zhi},
	month = aug,
	year = {2023},
	pages = {385--404},
}

@article{hong_more_2021,
	title = {More {Diverse} {Means} {Better}: {Multimodal} {Deep} {Learning} {Meets} {Remote}-{Sensing} {Imagery} {Classification}},
	volume = {59},
	issn = {1558-0644},
	shorttitle = {More {Diverse} {Means} {Better}},
	url = {https://ieeexplore.ieee.org/document/9174822/},
	doi = {10.1109/TGRS.2020.3016820},
	number = {5},
	urldate = {2026-02-03},
	journal = {IEEE Transactions on Geoscience and Remote Sensing},
	author = {Hong, Danfeng and Gao, Lianru and Yokoya, Naoto and Yao, Jing and Chanussot, Jocelyn and Du, Qian and Zhang, Bing},
	month = may,
	year = {2021},
	pages = {4340--4354},
}

@article{chen_novel_2024,
	title = {A {Novel} {Approach} to {Incomplete} {Multimodal} {Learning} for {Remote} {Sensing} {Data} {Fusion}},
	volume = {62},
	issn = {1558-0644},
	url = {https://ieeexplore.ieee.org/document/10496710/},
	doi = {10.1109/TGRS.2024.3387837},
	urldate = {2026-02-03},
	journal = {IEEE Transactions on Geoscience and Remote Sensing},
	author = {Chen, Yuxing and Zhao, Maofan and Bruzzone, Lorenzo},
	year = {2024},
	pages = {1--14},
}

@article{li_semantic_2026,
	title = {Semantic segmentation with scale alignment and contextual information fusion for multimodal remote sensing images},
	volume = {126},
	issn = {15662535},
	url = {https://linkinghub.elsevier.com/retrieve/pii/S1566253525007432},
	doi = {10.1016/j.inffus.2025.103671},
	language = {en},
	urldate = {2026-02-03},
	journal = {Information Fusion},
	author = {Li, Jiayuan and Wang, Zhen and Xu, Nan and You, Zhuhong},
	month = feb,
	year = {2026},
	pages = {103671},
}

@article{liu_cross-city_2024,
	title = {Cross-{City} {Semantic} {Segmentation} ({C2Seg}) in {Multimodal} {Remote} {Sensing}: {Outcome} of the 2023 {IEEE} {WHISPERS} {C2Seg} {Challenge}},
	volume = {17},
	copyright = {https://creativecommons.org/licenses/by-nc-nd/4.0/},
	issn = {1939-1404, 2151-1535},
	shorttitle = {Cross-{City} {Semantic} {Segmentation} ({C2Seg}) in {Multimodal} {Remote} {Sensing}},
	url = {https://ieeexplore.ieee.org/document/10517985/},
	doi = {10.1109/JSTARS.2024.3388464},
	language = {en},
	urldate = {2026-02-03},
	journal = {IEEE Journal of Selected Topics in Applied Earth Observations and Remote Sensing},
	author = {Liu, Yuheng and Wang, Ye and Zhang, Yifan and Mei, Shaohui and Zou, Jiaqi and Li, Zhuohong and Lu, Fangxiao and He, Wei and Zhang, Hongyan and Zhao, Huilin and Chen, Chuan and Xia, Cong and Li, Hao and Vivone, Gemine and Hänsch, Ronny and Taskin, Gulsen and Yao, Jing and Qin, A. K. and Zhang, Bing and Chanussot, Jocelyn and Hong, Danfeng},
	year = {2024},
	pages = {8851--8862},
}

@article{hosseinpour_cmgfnet_2022,
	title = {{CMGFNet}: {A} deep cross-modal gated fusion network for building extraction from very high-resolution remote sensing images},
	volume = {184},
	issn = {09242716},
	shorttitle = {{CMGFNet}},
	url = {https://linkinghub.elsevier.com/retrieve/pii/S0924271621003294},
	doi = {10.1016/j.isprsjprs.2021.12.007},
	language = {en},
	urldate = {2026-02-03},
	journal = {ISPRS Journal of Photogrammetry and Remote Sensing},
	author = {Hosseinpour, Hamidreza and Samadzadegan, Farhad and Javan, Farzaneh Dadrass},
	month = feb,
	year = {2022},
	pages = {96--115},
}

@article{ma_crossmodal_2022,
	title = {A {Crossmodal} {Multiscale} {Fusion} {Network} for {Semantic} {Segmentation} of {Remote} {Sensing} {Data}},
	volume = {15},
	copyright = {https://creativecommons.org/licenses/by/4.0/legalcode},
	issn = {1939-1404, 2151-1535},
	url = {https://ieeexplore.ieee.org/document/9749821/},
	doi = {10.1109/JSTARS.2022.3165005},
	language = {en},
	urldate = {2026-02-03},
	journal = {IEEE Journal of Selected Topics in Applied Earth Observations and Remote Sensing},
	author = {Ma, Xianping and Zhang, Xiaokang and Pun, Man-On},
	year = {2022},
	pages = {3463--3474},
}

@article{baltrusaitis_multimodal_2019,
	title = {Multimodal {Machine} {Learning}: {A} {Survey} and {Taxonomy}},
	volume = {41},
	issn = {1939-3539},
	shorttitle = {Multimodal {Machine} {Learning}},
	url = {https://ieeexplore.ieee.org/document/8269806/},
	doi = {10.1109/TPAMI.2018.2798607},
	number = {2},
	urldate = {2026-02-03},
	journal = {IEEE Transactions on Pattern Analysis and Machine Intelligence},
	author = {Baltrušaitis, Tadas and Ahuja, Chaitanya and Morency, Louis-Philippe},
	month = feb,
	year = {2019},
	pages = {423--443},
}

@inproceedings{wang_self-supervised_2022,
	address = {Kuala Lumpur, Malaysia},
	title = {Self-{Supervised} {Vision} {Transformers} for {Joint} {SAR}-{Optical} {Representation} {Learning}},
	copyright = {https://doi.org/10.15223/policy-029},
	isbn = {978-1-6654-2792-0},
	url = {https://ieeexplore.ieee.org/document/9883983/},
	doi = {10.1109/IGARSS46834.2022.9883983},
	language = {en},
	urldate = {2026-02-03},
	booktitle = {{IGARSS} 2022 - 2022 {IEEE} {International} {Geoscience} and {Remote} {Sensing} {Symposium}},
	publisher = {IEEE},
	author = {Wang, Yi and Albrecht, Conrad M and Zhu, Xiao Xiang},
	month = jul,
	year = {2022},
	pages = {139--142},
}

@article{wei_improving_nodate,
	title = {Improving {Multimodal} {Learning} via {Imbalanced} {Learning}},
	language = {en},
	author = {Wei, Shicai and Luo, Chunbo and Luo, Yang},
}

@article{zheng_reducing_nodate,
	title = {Reducing {Unimodal} {Bias} in {Multi}-{Modal} {Semantic} {Segmentation} with {Multi}-{Scale} {Functional} {Entropy} {Regularization}},
	language = {en},
	author = {Zheng, Xu and Lyu, Yuanhuiyi and Jiang, Lutao and Paudel, Danda Pani and Gool, Luc Van and Hu, Xuming},
}

@article{sun_fully_2018,
	title = {Fully {Convolutional} {Networks} for {Semantic} {Segmentation} of {Very} {High} {Resolution} {Remotely} {Sensed} {Images} {Combined} {With} {DSM}},
	volume = {15},
	issn = {1558-0571},
	url = {https://ieeexplore.ieee.org/document/8281008/},
	doi = {10.1109/LGRS.2018.2795531},
	number = {3},
	urldate = {2026-02-03},
	journal = {IEEE Geoscience and Remote Sensing Letters},
	author = {Sun, Weiwei and Wang, Ruisheng},
	month = mar,
	year = {2018},
	pages = {474--478},
}

@article{parmehr_automatic_2014,
	title = {Automatic registration of optical imagery with {3D} {LiDAR} data using statistical similarity},
	volume = {88},
	issn = {09242716},
	url = {https://linkinghub.elsevier.com/retrieve/pii/S0924271613002773},
	doi = {10.1016/j.isprsjprs.2013.11.015},
	language = {en},
	urldate = {2026-02-03},
	journal = {ISPRS Journal of Photogrammetry and Remote Sensing},
	author = {Parmehr, Ebadat G. and Fraser, Clive S. and Zhang, Chunsun and Leach, Joseph},
	month = feb,
	year = {2014},
	pages = {28--40},
}

@article{zheng_gather--guide_2022,
	title = {A {Gather}-to-{Guide} {Network} for {Remote} {Sensing} {Semantic} {Segmentation} of {RGB} and {Auxiliary} {Image}},
	volume = {60},
	copyright = {https://ieeexplore.ieee.org/Xplorehelp/downloads/license-information/IEEE.html},
	issn = {0196-2892, 1558-0644},
	url = {https://ieeexplore.ieee.org/document/9519842/},
	doi = {10.1109/TGRS.2021.3103517},
	language = {en},
	urldate = {2026-02-03},
	journal = {IEEE Transactions on Geoscience and Remote Sensing},
	author = {Zheng, Xianwei and Wu, Xiujie and Huan, Linxi and He, Wei and Zhang, Hongyan},
	year = {2022},
	pages = {1--15},
}

@article{yang_attention-fused_2021,
	title = {An attention-fused network for semantic segmentation of very-high-resolution remote sensing imagery},
	volume = {177},
	issn = {09242716},
	url = {https://linkinghub.elsevier.com/retrieve/pii/S0924271621001295},
	doi = {10.1016/j.isprsjprs.2021.05.004},
	language = {en},
	urldate = {2026-02-03},
	journal = {ISPRS Journal of Photogrammetry and Remote Sensing},
	author = {Yang, Xuan and Li, Shanshan and Chen, Zhengchao and Chanussot, Jocelyn and Jia, Xiuping and Zhang, Bing and Li, Baipeng and Chen, Pan},
	month = jul,
	year = {2021},
	pages = {238--262},
}

@article{maggiori_high-resolution_2017,
	title = {High-{Resolution} {Aerial} {Image} {Labeling} {With} {Convolutional} {Neural} {Networks}},
	volume = {55},
	issn = {1558-0644},
	url = {https://ieeexplore.ieee.org/document/8024178/},
	doi = {10.1109/TGRS.2017.2740362},
	number = {12},
	urldate = {2026-02-02},
	journal = {IEEE Transactions on Geoscience and Remote Sensing},
	author = {Maggiori, Emmanuel and Tarabalka, Yuliya and Charpiat, Guillaume and Alliez, Pierre},
	month = dec,
	year = {2017},
	pages = {7092--7103},
}

@article{li_deep_2022,
	title = {Deep learning in multimodal remote sensing data fusion: {A} comprehensive review},
	volume = {112},
	issn = {15698432},
	shorttitle = {Deep learning in multimodal remote sensing data fusion},
	url = {https://linkinghub.elsevier.com/retrieve/pii/S1569843222001248},
	doi = {10.1016/j.jag.2022.102926},
	language = {en},
	urldate = {2026-02-02},
	journal = {International Journal of Applied Earth Observation and Geoinformation},
	author = {Li, Jiaxin and Hong, Danfeng and Gao, Lianru and Yao, Jing and Zheng, Ke and Zhang, Bing and Chanussot, Jocelyn},
	month = aug,
	year = {2022},
	pages = {102926},
}

@article{schmitt_data_2016,
	title = {Data {Fusion} and {Remote} {Sensing}: {An} ever-growing relationship},
	volume = {4},
	issn = {2168-6831},
	shorttitle = {Data {Fusion} and {Remote} {Sensing}},
	url = {https://ieeexplore.ieee.org/document/7740215/},
	doi = {10.1109/MGRS.2016.2561021},
	number = {4},
	urldate = {2026-02-02},
	journal = {IEEE Geoscience and Remote Sensing Magazine},
	author = {Schmitt, Michael and Zhu, Xiao Xiang},
	month = dec,
	year = {2016},
	pages = {6--23},
}

@article{bai_domain_2022,
	title = {Domain {Adaptation} for {Remote} {Sensing} {Image} {Semantic} {Segmentation}: {An} {Integrated} {Approach} of {Contrastive} {Learning} and {Adversarial} {Learning}},
	volume = {60},
	issn = {1558-0644},
	shorttitle = {Domain {Adaptation} for {Remote} {Sensing} {Image} {Semantic} {Segmentation}},
	url = {https://ieeexplore.ieee.org/document/9857935/},
	doi = {10.1109/TGRS.2022.3198972},
	urldate = {2026-02-02},
	journal = {IEEE Transactions on Geoscience and Remote Sensing},
	author = {Bai, Lubin and Du, Shihong and Zhang, Xiuyuan and Wang, Haoyu and Liu, Bo and Ouyang, Song},
	year = {2022},
	pages = {1--13},
}

@article{zhang_object-oriented_2014,
	title = {Object-{Oriented} {Shadow} {Detection} and {Removal} {From} {Urban} {High}-{Resolution} {Remote} {Sensing} {Images}},
	volume = {52},
	issn = {1558-0644},
	url = {https://ieeexplore.ieee.org/document/6779607/},
	doi = {10.1109/TGRS.2014.2306233},
	number = {11},
	urldate = {2026-02-02},
	journal = {IEEE Transactions on Geoscience and Remote Sensing},
	author = {Zhang, Hongya and Sun, Kaimin and Li, Wenzhuo},
	month = nov,
	year = {2014},
	pages = {6972--6982},
}

@article{liu_semantic_2018,
	title = {Semantic labeling in very high resolution images via a self-cascaded convolutional neural network},
	volume = {145},
	issn = {09242716},
	url = {https://linkinghub.elsevier.com/retrieve/pii/S0924271617303854},
	doi = {10.1016/j.isprsjprs.2017.12.007},
	language = {en},
	urldate = {2026-02-02},
	journal = {ISPRS Journal of Photogrammetry and Remote Sensing},
	author = {Liu, Yongcheng and Fan, Bin and Wang, Lingfeng and Bai, Jun and Xiang, Shiming and Pan, Chunhong},
	month = nov,
	year = {2018},
	pages = {78--95},
}

@article{volpi_dense_2017,
	title = {Dense {Semantic} {Labeling} of {Subdecimeter} {Resolution} {Images} {With} {Convolutional} {Neural} {Networks}},
	volume = {55},
	issn = {1558-0644},
	url = {https://ieeexplore.ieee.org/document/7725499/},
	doi = {10.1109/TGRS.2016.2616585},
	number = {2},
	urldate = {2026-02-02},
	journal = {IEEE Transactions on Geoscience and Remote Sensing},
	author = {Volpi, Michele and Tuia, Devis},
	month = feb,
	year = {2017},
	pages = {881--893},
}

@article{zheng_building_2021,
	title = {Building damage assessment for rapid disaster response with a deep object-based semantic change detection framework: {From} natural disasters to man-made disasters},
	volume = {265},
	issn = {00344257},
	shorttitle = {Building damage assessment for rapid disaster response with a deep object-based semantic change detection framework},
	url = {https://linkinghub.elsevier.com/retrieve/pii/S0034425721003564},
	doi = {10.1016/j.rse.2021.112636},
	language = {en},
	urldate = {2026-02-02},
	journal = {Remote Sensing of Environment},
	author = {Zheng, Zhuo and Zhong, Yanfei and Wang, Junjue and Ma, Ailong and Zhang, Liangpei},
	month = nov,
	year = {2021},
	pages = {112636},
}

@article{turkoglu_crop_2021,
	title = {Crop mapping from image time series: {Deep} learning with multi-scale label hierarchies},
	volume = {264},
	issn = {00344257},
	shorttitle = {Crop mapping from image time series},
	url = {https://linkinghub.elsevier.com/retrieve/pii/S0034425721003230},
	doi = {10.1016/j.rse.2021.112603},
	language = {en},
	urldate = {2026-02-02},
	journal = {Remote Sensing of Environment},
	author = {Turkoglu, Mehmet Ozgur and D'Aronco, Stefano and Perich, Gregor and Liebisch, Frank and Streit, Constantin and Schindler, Konrad and Wegner, Jan Dirk},
	month = oct,
	year = {2021},
	pages = {112603},
}

@article{chen_adaptive_2021,
	title = {Adaptive {Effective} {Receptive} {Field} {Convolution} for {Semantic} {Segmentation} of {VHR} {Remote} {Sensing} {Images}},
	volume = {59},
	issn = {1558-0644},
	url = {https://ieeexplore.ieee.org/document/9147012/},
	doi = {10.1109/TGRS.2020.3009143},
	number = {4},
	urldate = {2026-02-02},
	journal = {IEEE Transactions on Geoscience and Remote Sensing},
	author = {Chen, Xi and Li, Zhiqiang and Jiang, Jie and Han, Zhen and Deng, Shiyi and Li, Zhihong and Fang, Tao and Huo, Hong and Li, Qingli and Liu, Min},
	month = apr,
	year = {2021},
	pages = {3532--3546},
}

@article{diakogiannis_resunet-_2020,
	title = {{ResUNet}-a: {A} deep learning framework for semantic segmentation of remotely sensed data},
	volume = {162},
	issn = {09242716},
	shorttitle = {{ResUNet}-a},
	url = {https://linkinghub.elsevier.com/retrieve/pii/S0924271620300149},
	doi = {10.1016/j.isprsjprs.2020.01.013},
	language = {en},
	urldate = {2026-02-02},
	journal = {ISPRS Journal of Photogrammetry and Remote Sensing},
	author = {Diakogiannis, Foivos I. and Waldner, François and Caccetta, Peter and Wu, Chen},
	month = apr,
	year = {2020},
	pages = {94--114},
}

@article{oquab_dinov2_nodate,
	title = {{DINOv2}: {Learning} {Robust} {Visual} {Features} without {Supervision}},
	language = {en},
	author = {Oquab, Maxime and Darcet, Timothée and Moutakanni, Théo and Vo, Huy V and Szafraniec, Marc and Khalidov, Vasil and Fernandez, Pierre and Haziza, Daniel and Massa, Francisco and El-Nouby, Alaaeldin and Assran, Mahmoud and Ballas, Nicolas and Galuba, Wojciech and Howes, Russell and Huang, Po-Yao and Li, Shang-Wen and Misra, Ishan and Rabbat, Michael and Sharma, Vasu and Synnaeve, Gabriel and Xu, Hu and Jegou, Hervé and Mairal, Julien and Labatut, Patrick and Joulin, Armand and Bojanowski, Piotr},
}

@article{yan_ringmo-sam_2023,
	title = {{RingMo}-{SAM}: {A} {Foundation} {Model} for {Segment} {Anything} in {Multimodal} {Remote}-{Sensing} {Images}},
	volume = {61},
	copyright = {https://ieeexplore.ieee.org/Xplorehelp/downloads/license-information/IEEE.html},
	issn = {0196-2892, 1558-0644},
	shorttitle = {{RingMo}-{SAM}},
	url = {https://ieeexplore.ieee.org/document/10315957/},
	doi = {10.1109/TGRS.2023.3332219},
	language = {en},
	urldate = {2026-01-22},
	journal = {IEEE Transactions on Geoscience and Remote Sensing},
	author = {Yan, Zhiyuan and Li, Junxi and Li, Xuexue and Zhou, Ruixue and Zhang, Wenkai and Feng, Yingchao and Diao, Wenhui and Fu, Kun and Sun, Xian},
	year = {2023},
	pages = {1--16},
}

@article{zou_adapting_2025,
	title = {Adapting {Vision} {Foundation} {Models} for {Robust} {Cloud} {Segmentation} in {Remote} {Sensing} {Images}},
	volume = {63},
	copyright = {https://ieeexplore.ieee.org/Xplorehelp/downloads/license-information/IEEE.html},
	issn = {0196-2892, 1558-0644},
	url = {https://ieeexplore.ieee.org/document/11121911/},
	doi = {10.1109/TGRS.2025.3597410},
	language = {en},
	urldate = {2026-01-21},
	journal = {IEEE Transactions on Geoscience and Remote Sensing},
	author = {Zou, Xuechao and Zhang, Shun and Li, Kai and Wang, Shiying and Xing, Junliang and Jin, Lei and Lang, Congyan and Tao, Pin},
	year = {2025},
	pages = {1--14},
}

@article{liu_transformer-based_2024,
	title = {A {Transformer}-based multi-modal fusion network for semantic segmentation of high-resolution remote sensing imagery},
	volume = {133},
	issn = {15698432},
	url = {https://linkinghub.elsevier.com/retrieve/pii/S1569843224004370},
	doi = {10.1016/j.jag.2024.104083},
	language = {en},
	urldate = {2026-01-20},
	journal = {International Journal of Applied Earth Observation and Geoinformation},
	author = {Liu, Yutong and Gao, Kun and Wang, Hong and Yang, Zhijia and Wang, Pengyu and Ji, Shijing and Huang, Yanjun and Zhu, Zhenyu and Zhao, Xiaobin},
	month = sep,
	year = {2024},
	pages = {104083},
}

@article{zhou_cegfnet_2022,
	title = {{CEGFNet}: {Common} {Extraction} and {Gate} {Fusion} {Network} for {Scene} {Parsing} of {Remote} {Sensing} {Images}},
	volume = {60},
	copyright = {https://ieeexplore.ieee.org/Xplorehelp/downloads/license-information/IEEE.html},
	issn = {0196-2892, 1558-0644},
	shorttitle = {{CEGFNet}},
	url = {https://ieeexplore.ieee.org/document/9538389/},
	doi = {10.1109/TGRS.2021.3109626},
	language = {en},
	urldate = {2026-01-20},
	journal = {IEEE Transactions on Geoscience and Remote Sensing},
	author = {Zhou, Wujie and Jin, Jianhui and Lei, Jingsheng and Hwang, Jenq-Neng},
	year = {2022},
	pages = {1--10},
}

@article{houlsby_parameter-efficient_nodate,
	title = {Parameter-{Efficient} {Transfer} {Learning} for {NLP}},
	language = {en},
	author = {Houlsby, Neil and Giurgiu, Andrei and Jastrzebski, Stanisław and Morrone, Bruna},
}

@misc{dosovitskiy_image_2021,
	title = {An {Image} is {Worth} 16x16 {Words}: {Transformers} for {Image} {Recognition} at {Scale}},
	shorttitle = {An {Image} is {Worth} 16x16 {Words}},
	url = {http://arxiv.org/abs/2010.11929},
	doi = {10.48550/arXiv.2010.11929},
	language = {en},
	urldate = {2025-12-20},
	publisher = {arXiv},
	author = {Dosovitskiy, Alexey and Beyer, Lucas and Kolesnikov, Alexander and Weissenborn, Dirk and Zhai, Xiaohua and Unterthiner, Thomas and Dehghani, Mostafa and Minderer, Matthias and Heigold, Georg and Gelly, Sylvain and Uszkoreit, Jakob and Houlsby, Neil},
	month = jun,
	year = {2021},
	note = {arXiv:2010.11929 [cs]},
}

@incollection{avidan_visual_2022,
	address = {Cham},
	title = {Visual {Prompt} {Tuning}},
	volume = {13693},
	isbn = {978-3-031-19826-7 978-3-031-19827-4},
	url = {https://link.springer.com/10.1007/978-3-031-19827-4_41},
	doi = {10.1007/978-3-031-19827-4_41},
	language = {en},
	urldate = {2025-12-20},
	booktitle = {Computer {Vision} – {ECCV} 2022},
	publisher = {Springer Nature Switzerland},
	author = {Jia, Menglin and Tang, Luming and Chen, Bor-Chun and Cardie, Claire and Belongie, Serge and Hariharan, Bharath and Lim, Ser-Nam},
	editor = {Avidan, Shai and Brostow, Gabriel and Cissé, Moustapha and Farinella, Giovanni Maria and Hassner, Tal},
	year = {2022},
	note = {Series Title: Lecture Notes in Computer Science},
	pages = {709--727},
}

@article{ma_deep_2019,
	title = {Deep learning in remote sensing applications: {A} meta-analysis and review},
	volume = {152},
	issn = {09242716},
	shorttitle = {Deep learning in remote sensing applications},
	url = {https://linkinghub.elsevier.com/retrieve/pii/S0924271619301108},
	doi = {10.1016/j.isprsjprs.2019.04.015},
	language = {en},
	urldate = {2025-12-20},
	journal = {ISPRS Journal of Photogrammetry and Remote Sensing},
	author = {Ma, Lei and Liu, Yu and Zhang, Xueliang and Ye, Yuanxin and Yin, Gaofei and Johnson, Brian Alan},
	month = jun,
	year = {2019},
	pages = {166--177},
}

@article{zhu_deep_2017,
	title = {Deep {Learning} in {Remote} {Sensing}: {A} {Comprehensive} {Review} and {List} of {Resources}},
	volume = {5},
	copyright = {https://ieeexplore.ieee.org/Xplorehelp/downloads/license-information/IEEE.html},
	issn = {2168-6831, 2473-2397, 2373-7468},
	shorttitle = {Deep {Learning} in {Remote} {Sensing}},
	url = {https://ieeexplore.ieee.org/document/8113128/},
	doi = {10.1109/MGRS.2017.2762307},
	language = {en},
	number = {4},
	urldate = {2025-12-20},
	journal = {IEEE Geoscience and Remote Sensing Magazine},
	author = {Zhu, Xiao Xiang and Tuia, Devis and Mou, Lichao and Xia, Gui-Song and Zhang, Liangpei and Xu, Feng and Fraundorfer, Friedrich},
	month = dec,
	year = {2017},
	pages = {8--36},
}

@article{hu_lora_2022,
	title = {{LORA}: {LOW}-{RANK} {ADAPTATION} {OF} {LARGE} {LAN}- {GUAGE} {MODELS}},
	language = {en},
	author = {Hu, Edward and Shen, Yelong and Wallis, Phillip and Allen-Zhu, Zeyuan and Li, Yuanzhi and Wang, Shean and Wang, Lu and Chen, Weizhu},
	year = {2022},
}

@inproceedings{guo_skysense_2024,
	address = {Seattle, WA, USA},
	title = {{SkySense}: {A} {Multi}-{Modal} {Remote} {Sensing} {Foundation} {Model} {Towards} {Universal} {Interpretation} for {Earth} {Observation} {Imagery}},
	copyright = {https://doi.org/10.15223/policy-029},
	isbn = {979-8-3503-5300-6},
	shorttitle = {{SkySense}},
	url = {https://ieeexplore.ieee.org/document/10655854/},
	doi = {10.1109/CVPR52733.2024.02613},
	language = {en},
	urldate = {2025-09-01},
	booktitle = {2024 {IEEE}/{CVF} {Conference} on {Computer} {Vision} and {Pattern} {Recognition} ({CVPR})},
	publisher = {IEEE},
	author = {Guo, Xin and Lao, Jiangwei and Dang, Bo and Zhang, Yingying and Yu, Lei and Ru, Lixiang and Zhong, Liheng and Huang, Ziyuan and Wu, Kang and Hu, Dingxiang and He, Huimei and Wang, Jian and Chen, Jingdong and Yang, Ming and Zhang, Yongjun and Li, Yansheng},
	month = jun,
	year = {2024},
	pages = {27662--27673},
}

@misc{xiao_unified_2018,
	title = {Unified {Perceptual} {Parsing} for {Scene} {Understanding}},
	url = {http://arxiv.org/abs/1807.10221},
	doi = {10.48550/arXiv.1807.10221},
	language = {en},
	urldate = {2025-08-17},
	publisher = {arXiv},
	author = {Xiao, Tete and Liu, Yingcheng and Zhou, Bolei and Jiang, Yuning and Sun, Jian},
	month = jul,
	year = {2018},
	note = {arXiv:1807.10221 [cs]},
}

@article{hu_airs_2024,
	title = {{AiRs}: {Adapter} in {Remote} {Sensing} for {Parameter}-{Efficient} {Transfer} {Learning}},
	volume = {62},
	copyright = {https://ieeexplore.ieee.org/Xplorehelp/downloads/license-information/IEEE.html},
	issn = {0196-2892, 1558-0644},
	shorttitle = {{AiRs}},
	url = {https://ieeexplore.ieee.org/document/10385180/},
	doi = {10.1109/TGRS.2024.3351889},
	language = {en},
	urldate = {2025-08-11},
	journal = {IEEE Transactions on Geoscience and Remote Sensing},
	author = {Hu, Leiyi and Yu, Hongfeng and Lu, Wanxuan and Yin, Dongshuo and Sun, Xian and Fu, Kun},
	year = {2024},
	pages = {1--18},
}

@article{ma_unified_2025,
	title = {A {Unified} {Framework} {With} {Multimodal} {Fine}-{Tuning} for {Remote} {Sensing} {Semantic} {Segmentation}},
	volume = {63},
	copyright = {https://ieeexplore.ieee.org/Xplorehelp/downloads/license-information/IEEE.html},
	issn = {0196-2892, 1558-0644},
	url = {https://ieeexplore.ieee.org/document/11063320/},
	doi = {10.1109/TGRS.2025.3585238},
	language = {en},
	urldate = {2025-08-04},
	journal = {IEEE Transactions on Geoscience and Remote Sensing},
	author = {Ma, Xianping and Zhang, Xiaokang and Pun, Man-On and Huang, Bo},
	year = {2025},
	pages = {1--15},
}

@article{hu_tea_2024,
	title = {{TEA}: {A} {Training}-{Efficient} {Adapting} {Framework} for {Tuning} {Foundation} {Models} in {Remote} {Sensing}},
	volume = {62},
	copyright = {https://ieeexplore.ieee.org/Xplorehelp/downloads/license-information/IEEE.html},
	issn = {0196-2892, 1558-0644},
	shorttitle = {{TEA}},
	url = {https://ieeexplore.ieee.org/document/10740309/},
	doi = {10.1109/tgrs.2024.3488731},
	language = {en},
	urldate = {2025-07-15},
	journal = {IEEE Transactions on Geoscience and Remote Sensing},
	publisher = {Institute of Electrical and Electronics Engineers (IEEE)},
	author = {Hu, Leiyi and Lu, Wanxuan and Yu, Hongfeng and Yin, Dongshuo and Sun, Xian and Fu, Kun},
	year = {2024},
	pages = {1--18},
}

@article{li_abcnet_2021,
	title = {{ABCNet}: {Attentive} bilateral contextual network for efficient semantic segmentation of {Fine}-{Resolution} remotely sensed imagery},
	volume = {181},
	issn = {09242716},
	shorttitle = {{ABCNet}},
	url = {https://linkinghub.elsevier.com/retrieve/pii/S0924271621002379},
	doi = {10.1016/j.isprsjprs.2021.09.005},
	language = {en},
	urldate = {2025-05-13},
	journal = {ISPRS Journal of Photogrammetry and Remote Sensing},
	author = {Li, Rui and Zheng, Shunyi and Zhang, Ce and Duan, Chenxi and Wang, Libo and Atkinson, Peter M.},
	month = nov,
	year = {2021},
	pages = {84--98},
}

@inproceedings{kirillov_segment_2023,
	address = {Paris, France},
	title = {Segment {Anything}},
	copyright = {https://doi.org/10.15223/policy-029},
	isbn = {979-8-3503-0718-4},
	url = {https://ieeexplore.ieee.org/document/10378323/},
	doi = {10.1109/ICCV51070.2023.00371},
	language = {en},
	urldate = {2025-05-12},
	booktitle = {2023 {IEEE}/{CVF} {International} {Conference} on {Computer} {Vision} ({ICCV})},
	publisher = {IEEE},
	author = {Kirillov, Alexander and Mintun, Eric and Ravi, Nikhila and Mao, Hanzi and Rolland, Chloe and Gustafson, Laura and Xiao, Tete and Whitehead, Spencer and Berg, Alexander C. and Lo, Wan-Yen and Dollár, Piotr and Girshick, Ross},
	month = oct,
	year = {2023},
	pages = {3992--4003},
}

@incollection{lai_fusenet_2017,
	address = {Cham},
	title = {{FuseNet}: {Incorporating} {Depth} into {Semantic} {Segmentation} via {Fusion}-{Based} {CNN} {Architecture}},
	volume = {10111},
	isbn = {978-3-319-54180-8 978-3-319-54181-5},
	shorttitle = {{FuseNet}},
	url = {https://link.springer.com/10.1007/978-3-319-54181-5_14},
	doi = {10.1007/978-3-319-54181-5_14},
	language = {en},
	urldate = {2025-05-12},
	booktitle = {Computer {Vision} – {ACCV} 2016},
	publisher = {Springer International Publishing},
	author = {Hazirbas, Caner and Ma, Lingni and Domokos, Csaba and Cremers, Daniel},
	editor = {Lai, Shang-Hong and Lepetit, Vincent and Nishino, Ko and Sato, Yoichi},
	year = {2017},
	note = {Series Title: Lecture Notes in Computer Science},
	pages = {213--228},
}

@article{ma_multilevel_2024,
	title = {A {Multilevel} {Multimodal} {Fusion} {Transformer} for {Remote} {Sensing} {Semantic} {Segmentation}},
	volume = {62},
	issn = {1558-0644},
	url = {https://ieeexplore.ieee.org/document/10458980/},
	doi = {10.1109/TGRS.2024.3373033},
	urldate = {2025-04-17},
	journal = {IEEE Transactions on Geoscience and Remote Sensing},
	author = {Ma, Xianping and Zhang, Xiaokang and Pun, Man-On and Liu, Ming},
	year = {2024},
	pages = {1--15},
}

@article{wang_unetformer_2022,
	title = {{UNetFormer}: {A} {UNet}-like transformer for efficient semantic segmentation of remote sensing urban scene imagery},
	volume = {190},
	issn = {09242716},
	shorttitle = {{UNetFormer}},
	url = {https://linkinghub.elsevier.com/retrieve/pii/S0924271622001654},
	doi = {10.1016/j.isprsjprs.2022.06.008},
	language = {en},
	urldate = {2025-04-13},
	journal = {ISPRS Journal of Photogrammetry and Remote Sensing},
	author = {Wang, Libo and Li, Rui and Zhang, Ce and Fang, Shenghui and Duan, Chenxi and Meng, Xiaoliang and Atkinson, Peter M.},
	month = aug,
	year = {2022},
	pages = {196--214},
}

@article{li_mcanet_2022,
	title = {{MCANet}: {A} joint semantic segmentation framework of optical and {SAR} images for land use classification},
	volume = {106},
	issn = {15698432},
	shorttitle = {{MCANet}},
	url = {https://linkinghub.elsevier.com/retrieve/pii/S0303243421003457},
	doi = {10.1016/j.jag.2021.102638},
	language = {en},
	urldate = {2025-04-11},
	journal = {International Journal of Applied Earth Observation and Geoinformation},
	author = {Li, Xue and Zhang, Guo and Cui, Hao and Hou, Shasha and Wang, Shunyao and Li, Xin and Chen, Yujia and Li, Zhijiang and Zhang, Li},
	month = feb,
	year = {2022},
	pages = {102638},
}

%







\end{document}